%% file: main.tex
\documentclass[review]{elsarticle} 

\usepackage{lineno}
\usepackage{algorithm}
\usepackage{algorithmic}
\usepackage{amssymb}
\usepackage{times}
\usepackage{soul}
\usepackage{url}
\usepackage[small]{caption}
\usepackage{graphicx}
\usepackage{amsmath}
\usepackage{booktabs}
\usepackage{natbib}
\usepackage{subfig}
\usepackage{float}
\usepackage{newtxtext, newtxmath} 
\usepackage{bm}
\usepackage{csquotes}
\usepackage{xcolor}
\DeclareMathOperator*{\argmax}{arg\,max}
\DeclareMathOperator*{\argmin}{arg\,min}
\newcommand{\counterfactual}{\ensuremath{%
  \Box\kern-1.5pt
  \raise1pt\hbox{$\mathord{\rightarrow}$}}}

\newcommand{\bpi}{\bm{\pi}}
\newcommand{\bs}{\bm{s}}
\newcommand{\bz}{\bm{z}}
\newcommand{\bzw}{\bm{z_w}}
\newcommand{\bpiz}{\bm{\pi}(\bm{z})}
\newcommand{\bpiza}{\bm{\pi}(\bm{z},a)}

\setcitestyle{authoryear, open={((},close={]}}

\modulolinenumbers[5]

\journal{Journal of Artificial Intelligence}

\begin{document}

\begin{frontmatter}

\title{Counterfactual State Explanations for Reinforcement Learning Agents via Generative Deep Learning}


\author[mainaddress]{Matthew L. Olson\corref{correspondingauthor}}
\ead{olsomatt@eecs.oregonstate.edu}
\author[mainaddress]{Roli Khanna}
\author[mainaddress]{Lawrence Neal}
\author[mainaddress]{Fuxin Li}
\author[mainaddress]{Weng-Keen Wong}
\cortext[correspondingauthor]{Corresponding author at: 1148 Kelley Engineering Center, Corvallis, OR 97331-5501, USA. Tel.: +1 541 737 3617.}
\address[mainaddress]{Oregon State University, OR, USA}






\begin{abstract}
Counterfactual explanations, which deal with ``why not?'' scenarios, can provide insightful explanations to an AI agent's behavior \citep{TimMillerSocialSciencePaper2019}. In this work, we focus on generating counterfactual explanations for deep reinforcement learning (RL) agents which operate in visual input environments like Atari. We introduce {\it counterfactual} {\it state} {\it explanations}, a novel example-based approach to counterfactual explanations based on generative deep learning. Specifically, a counterfactual state illustrates what minimal change is needed to an Atari game image such that the agent chooses a different action. We also evaluate the effectiveness of counterfactual states on human participants who are not machine learning experts. Our first user study investigates if humans can discern if the counterfactual state explanations are produced by the actual game or produced by a generative deep learning approach. Our second user study investigates if counterfactual state explanations can help non-expert participants identify a flawed agent; we compare against a baseline approach based on a nearest neighbor explanation which uses images from the actual game.
Our results indicate that counterfactual state explanations have sufficient fidelity to the actual game images to enable non-experts to more effectively identify a flawed RL agent compared to the nearest neighbor baseline and to having no explanation at all.
\end{abstract}


\end{frontmatter}

\section{Introduction}
\input{intro.tex}

\section{Related Work}
\input{related_work.tex}

\section{Methodology: A Generative Deep Learning Model for Counterfactual States}

\input{methods.tex}

\section{Methodology: User Studies}
\input{user_study.tex}

\section{Results}

\input{results_imgs.tex}

\input{results_user_study.tex}
\section{Discussion}
\input{discussion.tex}

\section{Conclusion}
\input{conclusion.tex}

\section{Acknowledgements}
This work was supported by DARPA under the grant N66001-17-2-4030. We would like to thank  Andrew Anderson, Margaret Burnett, Jonathan Dodge, Alan Fern, Stefan Lee, Neale Ratzlaff, and Janet Schmidt for their expertise and helpful comments.

\bibliography{biblio}

\clearpage
\appendix
\input{appendix.tex}

\end{document}

%% file: intro.tex
Despite the impressive advances made by deep reinforcement learning (RL) agents, their decision-making process is challenging for humans to understand. This limitation is a serious concern for settings in which trust and reliability are critical, and deploying RL agents in these settings requires ensuring that they are making decisions for the right reasons. To solve this problem, researchers are developing techniques to provide human-understandable answers to explanatory questions of the agent's decision-making.

Explanatory questions can be classified into three types \citep{TimMillerSocialSciencePaper2019,Pearl18}: ``What?'' (Associative reasoning), ``How?'' (Interventionist reasoning) and ``Why?'' (Counterfactual reasoning). Of the three types, "Why?" questions are the most challenging as it requires counterfactual reasoning \citep{lewis2013counterfactuals, wachter2017counterfactual}, which involves reasoning about alternate outcomes that have not happened; counterfactual reasoning in turn requires both associative and interventionist reasoning \citep{TimMillerSocialSciencePaper2019}. In our work, we present a counterfactual explanation method to tackle the "Why?" question in Miller's classification. More specifically, we answer the "Why not?" question by using a deep generative model that can visually change the current state to produce alternate outcomes.

\begin{figure*}[t]
    \centering
    {{\includegraphics[width=.4\linewidth]{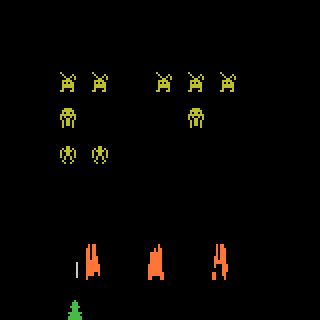} }}%
    \qquad
    {{\includegraphics[width=.4\linewidth]{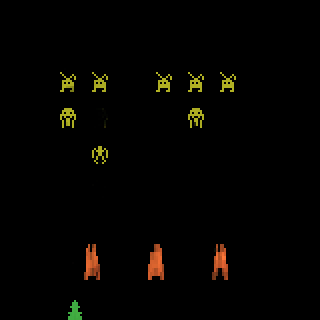} }}%
    \caption{A counterfactual example in the game of Space Invaders that demonstrates an agent's action changing by the removal of an enemy.
    \textbf{Left}: The game state in which an agent takes action $\text{``move left and shoot''}$.
    \label{fig:example}%
    \textbf{Right}: The counterfactual state where the agent will take the action $\text{``move right''}$.
    }
\end{figure*}

Underlying a RL agent is the mathematical framework of a Markov Decision Process (MDP) \citep{puterman94}, which models an agent making a sequence of decisions as it interacts with a stochastic environment. In the notation to follow in this section and in the rest of the manuscript, vectors, matrices and sets are in boldface while scalars are not. Formally, a MDP is a tuple $(\bm{S},\bm{\mathcal{A}},T,R,\gamma)$, where $\bm{S}$ is a set of states, $\bm{\mathcal{A}}$ is a set of actions, $T(\bm{s'},\bm{s},a)$ is a transition function capturing the probability of moving from state $\bm{s}$ to $\bm{s'}$ when action $a$ is performed in state $\bm{s}$, $R(\bm{s},a)$ is a reward function returning a reward for being in state $\bm{s}$ and performing action $a$ and $\gamma$ is called a discount factor (where $0 \leq \gamma \leq 1)$ which weights the importance of future rewards. 

Using the MDP framework, we introduce the concept of a counterfactual state as a counterfactual explanation\footnote{An early version of this work appeared in \citet{Olson19}.}.
More precisely, for an agent in state $\bm{s}$ performing action $a$ according to its learned policy, a counterfactual state $\bm{s'}$ is a state that involves a minimal change to $\bm{s}$ such that the agent's policy chooses action $a'$ instead of $a$. 
For example, a counterfactual state can be seen in Figure \ref{fig:example} for the video game Space Invaders \citep{brockman2016openai}. In this game, an agent exchanges fire with approaching enemies while taking cover underneath three barriers. 

Our approach is intended for deep RL agents that operate in visual input environments such as Atari. The main role of deep learning in these environments is to learn a lower dimensional representation of the state that captures the salient aspects needed to learn a successful policy. Our approach investigates how changes to the state cause the agent to choose a different action. As such, we do not focus on explaining the long term, sequential decision-making effects of following a learned policy, though this is a direction of interest for future work.

Our end goal is a tool for acceptance testing for end users of a deep RL agent. We envision counterfactual states being used in a replay environment in which a human user observes the agent as it executes its learned policy. At key frames in the replay, the user can ask the agent to generate counterfactual states which help the user determine if the agent has captured relevant aspects of the visual input for its decision making.

Our approach relies on a novel deep generative architecture to create counterfactual states. Past work on counterfactuals in visual input environments has relied on other techniques like part-swapping with a distractor image \citep{goyal_countervis_2019} or region in-filling \citep{chang2019explaining} to create counterfactual explanations. In contrast, our approach is more flexible in that it can generate entire counterfactual states images on demand by moving through the deep network's latent space.

We investigate the following research questions in this work:
\begin{enumerate}
    \item {\bf RQ1}: Can deep generative models produce high-fidelity counterfactual states that appear as if they are generated by the Atari game?
    \item {\bf RQ2}: Can counterfactual states help human users, who are non-experts in machine learning, understand enough of an agent's decision making to identify a flawed agent? 
    \item {\bf RQ3}: Can counterfactual states be more effective for helping users understand an agent's decision-making process than a nearest neighbor baseline technique?
\end{enumerate}

Our contributions are thus twofold. First, we introduce a new deep generative approach to generate counterfactual states to provide insight as to a RL agent's decision making. Second, we present results of user studies that investigate these research questions. Our results indicate that counterfactual states explanations are indeed useful. In our studies, they have sufficient fidelity to aid non-experts in identifying flawed RL agents.

%% file: related_work.tex

\subsection{Explainable Artificial Intelligence}


The literature on explainable AI is vast and we briefly summarize only the most directly related work.  Much of the past work on explaining machine learning has focused on explaining what features or regions of visual input were important for a prediction / action. A large class of approaches of this type fall under saliency map techniques, which use properties of the gradient to estimate the effect of pixels on the output (e.g. \citep{simonyan2013deep,springenberg2014striving,zeiler2014visualizing,selvaraju2017grad,fong2017interpretable,shrikumar2017learning,dabkowski2017real,Sundararajan17,zhang2018top,pmlr-v80-greydanus18a,qi2019visualizing}). Recent work, however, has found some saliency map techniques to be problematic. For instance, \citet{Adebayo18} found that some saliency map techniques still produced the same results even if the model parameters or the data labels were randomized. In addition, \citet{Atrey2020Exploratory} used counterfactual reasoning to evaluate if saliency maps were true explanations of an RL agent's behavior. Their findings indicated a negative result -- namely that saliency maps, by themselves, could lead to incorrect inferences by humans and should not be used as an explanation of an agent's behavior. Other explanation techniques include extracting a simpler interpretable model from a more complex model \citep{Craven95}, using locally interpretable models (e.g. \citep{Ribeiro18,Ribeiro16}), generating plots from Generalized Additive Models with pairwise interaction terms \citep{caruana15} and using influence functions to determine which training data instances most affect the prediction \citep{koh17}.




These methods, however, do not specifically identify changes in the current data instance that would result in a different outcome (or classification). These changes are a key part of the counterfactual reasoning needed to answer a "Why?" or "Why Not?" question. One of the first methods to do so was the Contrastive Explanations Method (CEM) \citep{Dhurandhar18}, which identified critical features or differences that would cause a data instance to be classified as another class. We found the hyperparameters for CEM to be difficult to tune to create high-fidelity counterfactuals for high-dimensional data like Atari images. As we will show in Section \ref{sec:realism}, CEM produced counterfactuals for Atari games that were filled with "snow" artifacts. CEM has also been extended to explain differences between policies in reinforcement learning \citep{vanderWaa18}. This approach focused on differences between trajectories in the environments rather than on the visual elements of a state, which is the focus of our work.

Two other recent approaches focused on producing counterfactuals for images. \citet{chang2019explaining} introduced the FIDO algorithm which generates counterfactuals for images by determining which regions, when filled in with values produced by a generative model, would most change the predicted class of the image.
The focus of the FIDO algorithm was on producing saliency maps and they used existing generative models for the infilling. In contrast, we develop a novel generative model to produce counterfactual state explanations; the goal of our method is to generate a realistic version of the entire counterfactual state (e.g. the whole Atari game frame image) in addition to producing difference highlights which are similar to saliency maps. Furthermore, \citet{chang2019explaining} did not evaluate their counterfactual explanations on human users while our user study results are one of our key contributions.

\citet{goyal_countervis_2019} generated counterfactual visual explanations for images by finding the minimal number of region swaps between the original image $\bm{I}$ with class $c$ and a distractor image $\bm{I'}$  with class $c'$ such that the class of $\bm{I}$ would change to $c'$. This method suffered from the problem that their counterfactual explanations could generate images with swapped regions that looked odd, e.g. due to pose misalignment between the two images. Their user study also focused on machine teaching, which is different from our focus of assessing agents for acceptance testing.

\subsection{Explainable Reinforcement Learning}


Past work on explaining RL has focused on explaining different aspects of the RL formulation. Techniques for explaining policies include explaining policies from Markov Decision Processes with logic-based templates \citep{Khan09}, state abstractions created through t-SNE embeddings \citep{Mnih15,zahavy2016graying}, human-interpretable predicates \citep{hayes2017improving}, high-level, domain-specific programming languages \citep{Verma18} and finite state machines for RNN policies \citep{koul2018learning}. \citet{Juozapaitis2019ExplainableRL} explained decisions made by RL agents by decomposing reward functions into simpler but semantically meaningful components. Finally, \citet{Mott2019TowardsIR} used an attention mechanism to identify relevant parts of the game environment for decision making. 

Another category of techniques for explaining RL used machine teaching to help end-users understand an agent's goals. \citet{Huang19} taught end-users about an agent's reward function using example trajectories chosen by an approximate-inference inverse RL algorithm. \citet{Lage19} investigated using both inverse RL and imitation learning to produce summaries of an agent's policy; their work highlighted the need for personalized summarization techniques as end-users varied in their preference of one technique over the other.

Other methods looked at summarizing an agent's behavior by presenting key moments of trajectories executed by a trained agent \citep{Amir18,huang2018,Sequeira20}. These key moments were intended to demonstrate an agent's capabilities, which could improve end-user trust. Key moments could be chosen by importance \citep{Amir18}, i.e. the largest difference in q-value for a given state \citep{Torrey13} or by critical states in which the q-value for one action was clearly superior to others \citep{huang2018}.  \citet{Sequeira20} explored interestingness based on the four dimensions of frequency, uncertainty, predictability and contradiction. For a summary, rather than presenting a single moment, they presented a sequence of states that varied according to a particular dimension.

These methods are all fundamentally different, yet complementary to our counterfactual approach of generating explanations. More specifically, our work can be used as an explanation technique to demonstrate an agent's proficiency once a key interaction moment has been chosen, such as by one of the aforementioned approaches.

\subsection{Generative Deep Learning}
As our counterfactuals are produced by a deep generative model, we briefly discuss related work on generative deep learning. Generative deep learning methods model the process that generates the data, thereby allowing never-before-seen data instances to be produced. Generative methods include auto-encoders \citep{ballard1987modular}, which encode an input feature vector into a lower-dimensional latent representation, and then decode that latent representation back to the original input space. Once the auto-encoder is trained, a common method to generate novel instances is to move about in the latent space and then decode the resulting latent space representation. However, these modifications in the latent space often result in unrealistic outputs \citep{bengio2013representation} due to "holes" in the learned latent space. This issue can be addressed by incorporating an additional loss function term that makes the latent representation match a pre-defined distribution \citep{VAE,Makhzani15,tolstikhin2018wasserstein}.

Another class of generative deep models are adversarial networks, which have gained increased attention due to their novel applications in modeling high-resolution data, especially generating faces that do not exist \citep{goodfellow2014generative}. Adversarial networks have been used to remove information predictive of the class label from a latent space. For example, Fader Networks \citep{lample2017fader} encoded an image of a flower to a lower dimensional latent representation that retained its shape and background, but did not contain information regarding its color (where color is the class label). The class label could then be combined with the latent representation to fully reconstruct the original data image, but crucially, the class label did not need to be the original one. This method could recreate many different versions of the same input that retained some properties, but had the characteristics relevant to the label changed. Thus, in this example, we can use Fader Networks to create a flower image with a specific shape and background, but with a different color from the original label.

%% file: methods.tex
\subsection{Counterfactual States}
The goal of this work is to shed some light into the decision making of a trained deep RL agent through counterfactual explanations. We are specifically interested in gaining some insight into what aspects of the visual input state $\bm{s}$ inform the choice of action $a$. Given a query state $\bm{s}$, we generate a \emph{counterfactual state} $\bm{s}'$ that minimally differs in some sense from $\bm{s}$, but results in the agent performing action $a'$ rather than action $a$. We refer to $a'$ as the \emph{counterfactual action}. 

\begin{figure}[htb]
\begin{center}
\includegraphics[width=0.75\columnwidth]{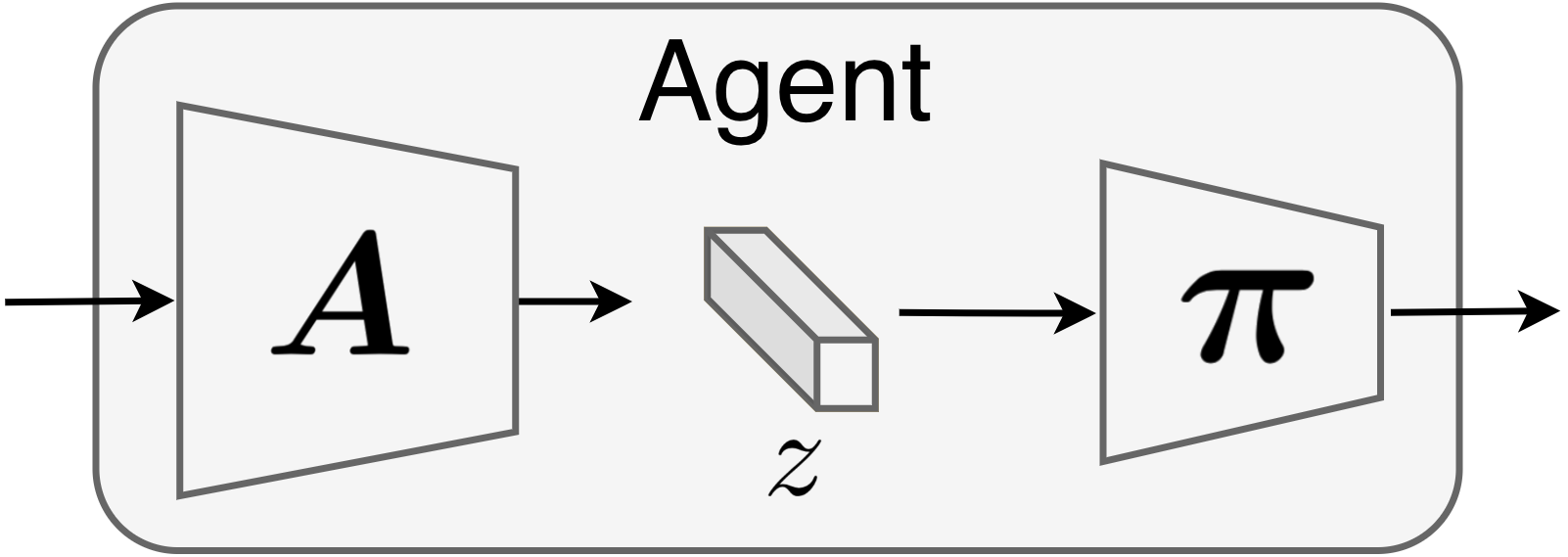}
\caption{The components of a pre-trained agent.}
\label{fig:train_architecture1}
\end{center}
\end{figure}

Our approach requires a trained deep RL agent to be given to us by an external party. We now describe this agent, illustrated in Figure \ref{fig:train_architecture1}. This agent has a learned policy represented by a deep neural network. We divide this policy network into two partitions of interest (Figure \ref{fig:train_architecture1}).  The first partition of the network layers, which we denote as $A$, takes a state $\bm{s}$ and maps it to a latent representation $\bm{z}=A(\bm{s})$. The vector $\bm{z}$ corresponds to the latent representation of $\bm{s}$ in the second to last fully connected layer in the network. The second partition of network layers, which we denote as $\bm{\pi}$, takes $\bm{z}$ and converts it to an action distribution $\bm{\pi}(\bm{z})$ i.e. a vector of probabilities for each action. Typically, $\bm{\pi}$ consists of a fully connected linear layer followed by a softmax. We use $\bm{\pi}(\bm{z},a)$ to refer to the probability of action $a$ in the action distribution $\bm{\pi}(\bm{z})$. We highlight the distinction in our Atari setting between a state $\bm{s}$, which is a raw Atari game image (also called a game frame), and the latent state $\bm{z}$ which is obtained from the second to last fully connected layer of the policy network. This latent layer, which we call $\bm{Z}$ is important in our diagnosis because it is used by the agent to inform its choice of actions. Our generative model is trained using a training dataset $\mathcal{X} = \{(\bm{s}_1,\bm{a}_1),\ldots,(\bm{s}_N,\bm{a}_N)\}$ of $N$ state-action pairs, where the action vectors $\bm{a}_i$ are action distributions obtained from the trained agent as it executes its learned policy. In summary, the agent\footnote{The agent may have other components such as the value function network. Our current work only uses the policy network, but we would like to apply similar ideas to the value function network.} can be viewed as the mapping $\bm{\pi}(A(\bm{s}))$.

Our approach to counterfactual explanations is to create counterfactual states using a deep generative model, which have been shown to produce realistic images \citep{radford2015unsupervised}. Our strategy is to encode the query state $\bm{s}$ to a latent representation. Then, from this latent representation, we move in the latent space $\bm{Z}$ in a direction that increases the probability of performing the counterfactual action $a'$. However, as previously noted by prior work, the latent space of a standard auto-encoder is filled with ``holes'' and counterfactual states generated from these holes would look unrealistic \citep{bengio2013representation}. To produce a latent space that is more amenable to creating representative outputs, we create a novel architecture that involves an adversarial auto-encoder \citep{Makhzani15} and a Wasserstein auto-encoder \citep{tolstikhin2018wasserstein}. Other approaches for navigating the latent space are possible, such as the methods presented by \citet{jahanian2019steerability} and \citet{Besserve2020Counterfactuals}, but these approaches do not specify an encoder, which is required in our framework to encode a query state $\bm{s}$ to a latent representation. 

\begin{figure}[t]
\begin{center}
\includegraphics[width=0.75\columnwidth]{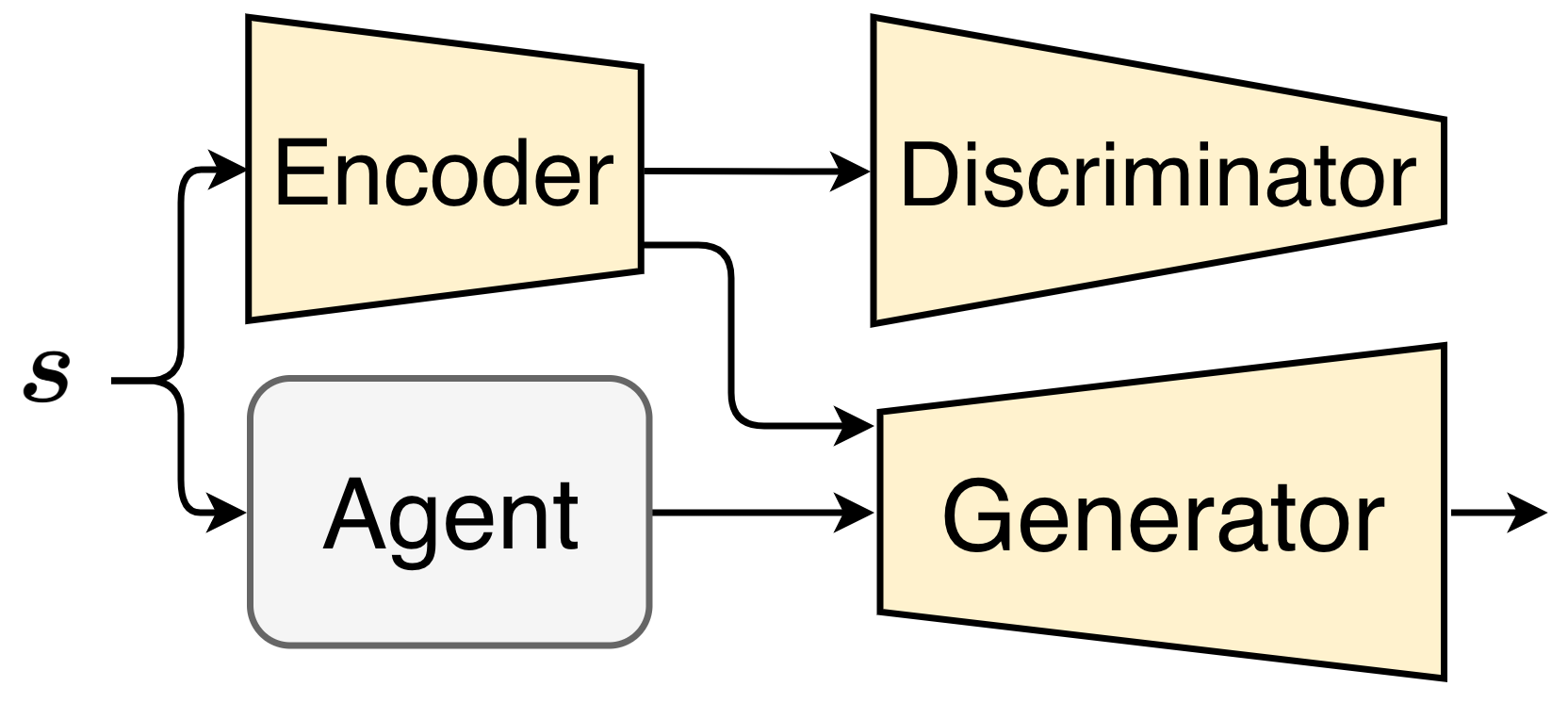}
\caption{An overview of our architecture, which consists of the encoder $E$, generator $G$, discriminator $D$ and pre-trained agent (grey).}
\label{fig:train_architecture2}
\end{center}
\end{figure}

\subsection{The Deep Network Architecture}
\label{sec:method_arch}
Figure \ref{fig:train_architecture2} depicts the architecture that we use during training. The RL agent is shaded gray to indicate that it has already been trained. First, we describe the Encoder ($E$), the Discriminator ($D$) and the Generator ($G$), which act together to produce counterfactual state images that vary depending on an input action distribution. Second, we describe the Wasserstein auto-encoder ($E_w, D_w$), which produces a new latent space based on the agent's latent space $\bm{Z}$; this new latent space enables perturbations within this space to produce meaningful counterfactual states. Each of these components contributes a loss term to the overall loss function used to train the network. 

\subsubsection{The Encoder, Discriminator and Generator}
\paragraph{\bf{Auto-encoder Loss}}
The encoder $E$ and generator $G$ act as an encoder-decoder pair. $E$ is a deep convolutional neural network that maps an input state $\bm{s}$ to a lower dimensional latent representation $E(\bm{s})$. We note that the Encoder $E$ is different from the encoder used by the agent's policy network and thus has a different latent space. $G$ is a deep convolutional generative neural network that creates an Atari image given its latent representation $E(\bm{s})$ and a policy vector $\bm{\pi}(\bm{z})$ (where $\bm{z} = A(s)$). 
The auto-encoding loss function of E and G is the mean squared error (MSE) function: 

\begin{equation}
L_{AE} = \frac{1}{|\mathcal{X}|} \sum_{(\bm{s},\bm{a}) \in \mathcal{X}} ||G(E(\bm{s}),\bm{\pi}(A(\bm{s}))) - \bm{s} ||^2_2
\label{eqn:autoencoderloss}
\end{equation}

To generate counterfactual states, we want to create a new image by changing the action distribution $\bm{\pi}(A(\bm{s}))$ to reflect the desired counterfactual action $a'$. However, in our experiments, we found that having only the loss function $L_{AE}$ by itself will cause $G$ to ignore $\bm{\pi}(A(\bm{s}))$ and use only $E(\bm{s})$; this behavior occurs because the loss function encourages reconstruction of $\bm{s}$ which can be achieved with only the encoding $E(\bm{s})$ and without $\bm{\pi}(A(\bm{s}))$. In order to make the Generator conditioned on the action distribution, we add an adversarial loss term using a discriminator $D$.

\paragraph{\bf{Discriminator Loss}}
In order to ensure that $\bm{\pi}(\bm{z})$ is not ignored, we cause the encoder to create an action-invariant representation $E(\bm{s})$. By action-invariant, we mean that the representation $E(\bm{s})$ no longer captures aspects of the state $\bm{s}$ that inform the choice of action. By doing so, adding $\bm{\pi}(\bm{z})$ as an input to $G$, along with $E(\bm{s})$, will provide the necessary information that will allow $G$ to recreate the effects of $\bm{\pi}$. In order to create an action-invariant representation, we perform adversarial training on the latent space, similar to the approach taken by \citet{lample2017fader}.
	
We thus add a discriminator $D$ that is trained to predict the full action distribution $\bm{\pi}(\bm{z})$ given $E(\bm{s})$. The action-invariant latent representation is learned by $E$ such that $D$ is unable to predict the true $\bm{\pi}(\bm{z})$ from our agent. As in Generative Adversarial Networks (GANs) \citep{goodfellow2014generative}, this setting corresponds to a two-player game where $D$ aims at maximizing its ability to identify the action distribution, and $E$ aims at preventing $D$ from being a good discriminator. The discriminator $D$ approximates $\bm{\pi}(\bm{z})$ given the encoded state $E(\bm{s})$, and is trained with MSE loss as shown below:

\begin{equation}
L_D =  \frac{1}{|\mathcal{X}|} \sum_{(\bm{s},\bm{a}) \in \mathcal{X}} ||D(E(\bm{s})) - \bm{\pi}(A(\bm{s})) ||^2_2
\label{eqn:discriminatorloss}
\end{equation}

\paragraph{\bf{Adversarial Loss}}
The objective of the encoder $E$ is now to learn a latent representation that optimizes two objectives. The first objective causes the generator to reconstruct the state $\bm{s}$ given $E(\bm{s})$ and $\bm{\pi}(A(\bm{s}))$, but the second objective causes the discriminator to be unable to predict $\bm{\pi}(A(\bm{s}))$ given $E(\bm{s})$. To accomplish this behavior in $D$, we want to maximize the entropy $H(D(E(\bm{s})))$, where $H(\bm{p}) = - \sum_i p_i log (p_i)$. Therefore, the adversarial loss can be written as:

\begin{equation}
L_{Adv} = \frac{\lambda}{|\mathcal{X}|} \sum_{(\bm{s},\bm{a}) \in \mathcal{X}} -H(D(E(\bm{s})))
\label{eqn:adversarialloss}
\end{equation}

The hyper-parameter $\lambda > 0$ weights the importance of this adversarial loss in the overall loss function. A larger $\lambda$ amplifies the importance of a high entropy $\bm{\pi}(\bm{z})$, which in turn reduces the amount of action-related information in $E(\bm{s})$ and if pushed to the extreme, results in the generator $G$ producing unrealistic game frames. On the other hand, small values of $\lambda$ lower $G$'s reliance on the input $\bm{\pi}(\bm{z})$, resulting in small changes to the game state when $\bm{\pi}(\bm{z})$ is modified. For analysis of the effects of varying $\lambda$, see \ref{lambda_tuning}.

\subsubsection{Wasserstein Autoencoder}

The counterfactual states require a notion of closeness between the query state $\bm{s}$ and the counterfactual state $\bm{s}'$. This notion of closeness can be measured in terms of distance in the agent's latent space $\bm{Z}$. We want to create a counterfactual state in the latent space $\bm{Z}$ because it directly influences the action distribution $\bm{\pi}$. We perform gradient descent in this feature space with respect to our target action $a'$ to produce a new $\bm{\pi}$ that has an increased probability of the counterfactual action $a'$. However, as previously mentioned, moving about in the  standard autoencoder's latent representation can result in unrealistic counterfactuals \citep{bengio2013representation}. To avoid this problem, we re-represent $\bm{Z}$ to a lower-dimensional manifold $\bm{Z_W}$ that is more compact and better-behaved for producing representative counterfactuals.

We use a Wasserstein auto-encoder (WAE) to learn a mapping function from the agent's original latent space to a well-behaved manifold \citep{tolstikhin2018wasserstein}. By using the concept of optimal transport, WAEs have shown that they can learn not just a low dimensional embedding, but also one where data points retain their concept of closeness in their original feature space where likely data points are close together.

The closeness-preserving nature of the WAE plays an important role when creating an action distribution vector $\bm{\pi}(\bm{z})$. In our counterfactual setting, we want to investigate the effect of performing action $a'$. However, we cannot simply convert $a'$ to an action distribution vector and assign a probability of 1 to the corresponding component in this vector as this approach could result in unrepresentative and low fidelity images. Instead, we follow a gradient in the $\bm{Z_W}$ space, which produces action distribution vectors that are more representative of those produced by the RL agent. This process, in turn, enables the Generator $G$ to produce more realistic images.

\begin{figure}[t]
\begin{center}
\includegraphics[width=0.6\columnwidth]{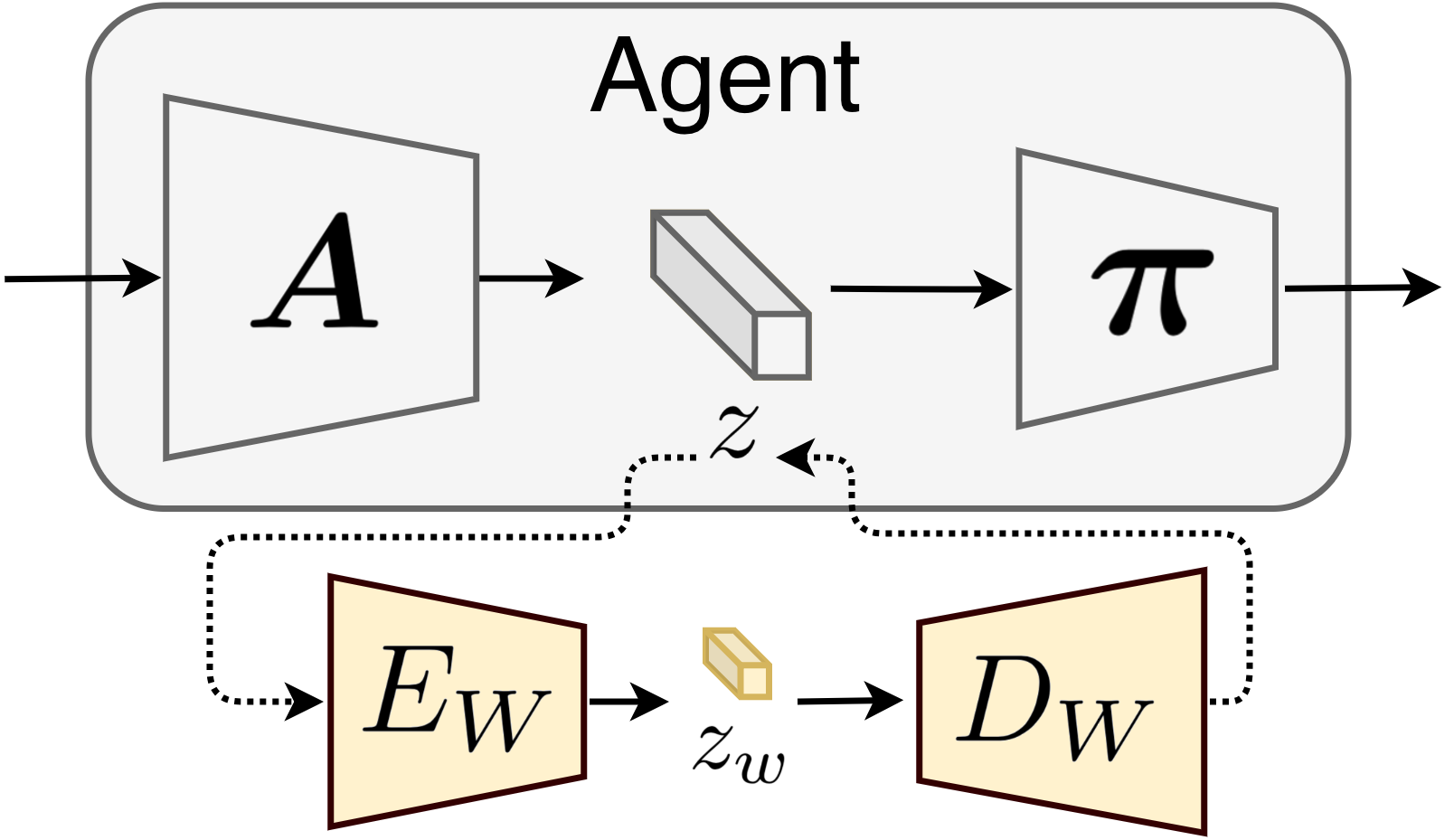}
\caption{The Wasserstein auto-encoder (shown as the pair $E_W$ and $D_W$) approximates the distribution of internal agent states $\bm{z}$.}
\label{fig:train_architecture3}
\end{center}
\end{figure}

We train a WAE, with encoder $E_W$ and decoder $D_W$, on data instances represented in the agent's latent space $\bm{Z}$ (see Figure~\ref{fig:train_architecture3}). We use MSE loss regularized by Maximum Mean Discrepancy (MMD):

\begin{equation}
L_{WAE} = \frac{1}{|S|} \sum_{\bm{s}} \left\Vert D_W(E_W(A(\bm{s}))) - A(\bm{s}) \right\Vert^2_2 \  +  MMD_k(D_W,E_W)
\label{eqn:wassersteinloss}
\end{equation}

where

\begin{equation}
MMD_k(D_Z,Q_Z) = \left\Vert \int_Z k(z,\cdot)dD_Z(z) - \int_Z k(z,\cdot)dE_Z(z) \right\Vert_{\mathcal{H}_k}
\end{equation}

Here $\mathcal{H}_k$ is a reproducing kernel Hilbert space, and in our work, an inverse multi-quadratic kernel is used \citep{tolstikhin2018wasserstein} .

\subsubsection{Training}
We let a pre-trained agent play the game with $\epsilon$-greedy exploration and train with the resulting dataset $\mathcal{X} = \{(\bm{s}_1, \bm{a}_1), \ldots, (\bm{s}_N,\bm{a}_N))$. We train with the overall loss function equal to $L = L_{AE} + L_D + L_{Adv} + L_{WAE}$. The loss function is minimized at each game time step with stochastic gradient descent using an ADAM optimizer \citep{ADAM}. 

\subsubsection{Loss function Clipping}
Generative models have been shown to have great difficulty in retaining small objects \citep{alvernaz2017autoencoder}. We follow \citep{kaiser2020model} by using loss clipping, which is defined as max($Loss,C$) for a constant $C$. This clipping is only applied to our auto-encoder and it is critical as many small gradients for each easy-to-predict background pixel outweigh the cost for mispredicting the hard-to-encode small objects. In our setting, we find that this loss clipping ensures the retention of small but key objects during auto-encoding and the creation of these objects when generating counterfactual states, such as the bullets in the Atari game Space Invaders.

\subsection{Generating Counterfactuals}
\label{sec:generating_counterfactuals}
Our goal is to generate counterfactual images that closely resemble real states of the game environment, but result in the agent taking action $a'$ instead of action $a$. In order to identify the necessary elements of the state that would need to be changed, we require that the generated counterfactual state $\bm{s'}$ is minimally changed from the original query state $\bm{s}$. Similar to \citet{neal2018open}, we formulate this process as an optimization:

\begin{align*}
    \text{minimize} &&
    ||E_w(A(\bm{s})) - \bm{z_w^*}||_2^2 \\
    \text{subject to} &&
    \argmax_{a \in \mathcal{A}}\ \  \bm{\pi}(D_W(\bm{z_w^*}), a) = a'
\end{align*}

where $\bm{s}$ is the given query state, $\mathcal{A}$ is the set of actions, and $\bm{z_w^*}$ is a latent point representing a possible internal state of the agent. This optimization can be relaxed as follows:

\begin{equation} \label{grad_descent}
\bm{z_w^*} = \argmin_{\bm{z_w}} \Bigg\{
||\bm{z_w} - E_w(A(\bm{s}))||_2^2
+
\log \left(1 - \bm{\pi}(D_W(\bm{z_w}), a' ) \right) \Bigg\}
\end{equation}

where $\bm{\pi}(\bm{z}, a)$ is the probability of the agent taking a discrete action $a$ on the counterfactual state representation $\bm{z}$. By minimizing the second term, we aim to increase the probability of taking action $a'$ and reduce the probability of taking all other actions. 

To generate a counterfactual state, we select a state from the training set, then encode the state to a Wasserstein latent point $\bm{z_w} = E_W(A(\bm{s}))$. We then minimize Equation \ref{grad_descent} through gradient descent with respect to $\bm{z_w}$ to find $\bm{z_w^*}$, then decode the latent point to create a new $\bm{\pi}(\bm{z})$ which is passed to the generator, along with $E(\bm{s})$ to create the counterfactual state $\bm{s'}$.

\subsection{Experimental Setup}
\label{sec:experimental_setup}

The pre-trained agent is a deep convolutional feed-forward network trained with Asynchronous Advantage Actor-Critic (A3C)  \citep{Mnih15} to maximize score in an Atari game. Games are played with a fixed frame-skip of 8 (7 for Space Invaders). The network that takes a set of 4 concatenated monochrome frames as input and is trained to maximize game score using the A3C algorithm.
We decompose the agent into two functions: $A(\bm{s})$ which takes as input 4 concatenated video frames and produces a 256-dimensional vector $\bm{z}$, and $\bm{\pi}(\bm{z})$ which outputs a distribution among actions. 
The frames are down sampled and cropped to 80x80, with normalized values [0,1]. This input is processed by 4 convolutional layers (each with 32 filters, kernel sizes of 3, strides of 2, and paddings of 1), followed by a fully connected layer, sized 256, and a last fully connected layer size $|\mathcal{A}| + 1$, where $|\mathcal{A}|$ is the action space size. We apply a softmax activation to the first $|\mathcal{A}|$ neurons to obtain $\pi(\bm{s})  = a$ and use the last neuron to predict the value, $V(\bm{s})$.

The A3C RL algorithm was trained with a learning rate of $\alpha = 10^{-4}$ , a discount factor of $\gamma = 0.99$, and computed loss on the policy using Generalized Advantage Estimation with $\lambda = 1.0$.
We find that convergence is more difficult with such a large frame skip, so each policy was trained asynchronously for a total of 50 million frames.

During training, we do not downscale or greyscale the game state. We pass in the current game time step as a 3 channels, RGB image. To generate the dataset $\mathcal{X}$, we set $\epsilon$ exploration value to $0.2$ and have the agent play for 25 million environment steps. 

\subsubsection{Network Details}
The encoder $E$ consists of 6 convolutional layers followed by 2 fully-connected layers with LeakyReLU activations and batch normalization. The output $E(\bm{s})$ is a 16-dimensional vector. For most of our agents, we find a value of $\lambda=50$ enforces a good trade-off between state reconstruction and reliance on $\bm{\pi}(\bm{z})$. The output of the network is referred to in the text as $E(s)$.

The generator $G$ consists of one fully-connected layer followed by 6 transposed convolutional layers, all with LeakyReLU activations and batch normalization. The encoded state $E(\bm{s})$ and the action distribution $\bm{\pi}(\bm{z})$ are fed to the first layer of the generator. 
Additionally, following the recommendation of \citet{lample2017fader}, $\bm{\pi}(\bm{z})$ is appended as an additional input channel to each subsequent layer, which ensures $G$ learns to depend on the values of $\bm{\pi}(\bm{z})$ for image creation when $\bm{\pi}(\bm{z})$ is modified during counterfactual generation.

The discriminator $D$ consists of two fully-connected layers followed by a softmax function, and outputs a distribution among actions with the same dimensionality as $\bm{\pi}(\bm{z})$.

The Wasserstein encoder $E_w$ consists of 3 fully-connected layers mapping $\bm{z}$ to a 128-dimensional vector $\bm{z}_w$, normalized such that $\left\Vert \bm{z_w} \right\Vert_2 = 1$. Each layer has the same dimensionality of 256, except the output of the 3rd layer which is 128. Additionally, the first two layers are followed by batch normalization and leaky ReLU with a leak of $0.2$.
The corresponding Wasserstein decoder $D_w$ is symmetric to $E_w$, with batch normalization and leaky ReLU after the first two layers and maps $\bm{z_w}$ back to $\bm{z}$.

\subsubsection{Training Details}
The encoder, generator, and discriminator are all trained through stochastic gradient descent using an Adam optimizer, with parameters $\alpha = 1^{-4}, \beta_{1} = 0, \beta_{2} = 0.9$. These networks were typically trained for 25 million game states to achieve high fidelity reconstructions, but we found even a tenth of the game states to be enough to produce meaningful counterfactual states. We set the max loss clipping constant $C=0.0001$, meaning if reconstructed  pixel (0-255) is within 2 values, its gradients are ignored. When training the agent, we use the current time step and previous 3 time steps concatenated to represent the state. For our generative model, we only use the current state. 

The Wasserstein Autoencoder was trained with Adam optimizers of the same learning rate $\alpha = 10^{-4}$ and with the default $\beta$ parameters. Training was performed for 15 million frames, upon which we found selecting actions from $\bpi(D_w(E_w(A(\bs))))$ consistently achieved the same average game score as the original agent.

All models are constructed and trained using PyTorch \citep{paszke2017automatic}.
For more information about our architecture and training parameters, our code can be accessed at: \textit{https://github.com/mattolson93/counterfactual-state-explanations/}

\subsubsection{Creating counterfactual state highlights}
A counterfactual state often contains small changes that are difficult to notice without careful inspection, so we mimic the saliency map generation process in \citet{pmlr-v80-greydanus18a} to highlight the difference between the original and counterfactual state. We take the absolute difference between the original state $\bm{s}$ and counterfactual state $\bm{s'}$ to create a counterfactual mask $\bm{m_c} = ||\bm{s} - \bm{s'}||_1$. For further clarity of the changes, we apply a Gaussian blur over the mask. Lastly, we set the blurred mask to a single color channel and combine this color mask with the original state to get the highlights. In our experiments, the highlights are in different colors for different games (e.g. blue for Space Invaders and red for Qbert) as we want colors that are a stark contrast from the color scheme of the game. 


%% file: user_study.tex

In general, evaluating explanations is a challenging problem, and counterfactual explanations are particularly difficult. A good counterfactual explanation helps humans understand why an agent performed a particular action. This human-based criterion is infeasible to capture with quantitative metrics. For instance, using the probability $\pi(\bm{s'},a')$ as a quantitative metric for a counterfactual state $\bm{s}'$ is misleading because this probability can be high for some Atari images that humans can immediately recognize as not generated by the game itself and also high for adversarial examples with imperceptible changes to the original state $\bm{s}$.  

Since evaluating counterfactuals requires human inspection, we designed two user studies. In the first user study, we evaluated the fidelity of our counterfactual states to the game.  By fidelity, we refer to how well the counterfactual images appear to be generated by the game itself rather than by a  generative deep learning model. In the second user study, we investigated if our counterfactual states could help humans understand enough of an agent's decision making so that they could perform a downstream task of identifying a flawed RL agent.

\subsection{User Study 1: Fidelity of Counterfactual States (RQ1)}
\begin{figure}[htb]
    \centering
    \includegraphics[width=.32\linewidth]{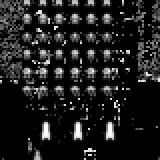} 
    \includegraphics[width=.32\linewidth]{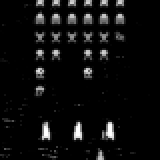} 
    \includegraphics[width=.32\linewidth]{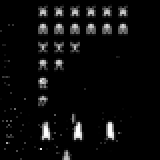} %
    \caption{Counterfactual states generated using the Contrastive Explanation Method with three choices of parameters on different states. Images are in black and white because the original CEM source code operates on direct input to the agent-- which are down-scaled, grey images.}
    
    \label{fig:CEM}%
\end{figure}

In order to evaluate the fidelity of our counterfactual states, we needed to create baseline methods for comparison. First, we experimented with using \textit{pertinent negatives} from the Contrastive Explanation Method (CEM) \citep{Dhurandhar18} as counterfactuals. These \textit{pertinent negatives} highlight absent features that would cause the agent to select an alternate action. We generated pertinent negatives from Atari states with pixels as features, and interpreted them as counterfactual states. We performed an extensive search over hyper-parameters to generate high-fidelity states, but found CEM very difficult to tune due to the high-dimensional nature of Atari images. The generated counterfactual states were either identical to the original query state or they had obvious ``snow'' artifacts as shown in Figure \ref{fig:CEM}, making them too low quality to serve as a reasonable baseline for our user study.

\begin{figure}[t]
    \centering
    \includegraphics[width=.32\linewidth]{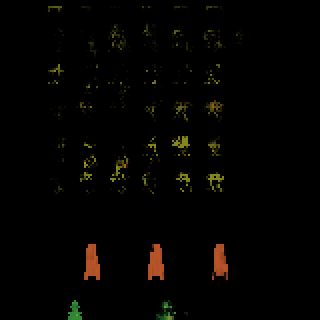} 
    \includegraphics[width=.32\linewidth]{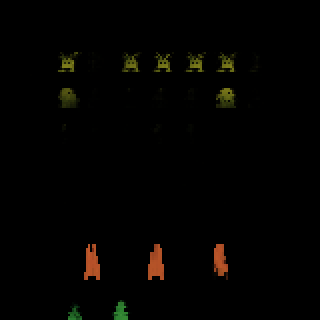} 
    \includegraphics[width=.32\linewidth]{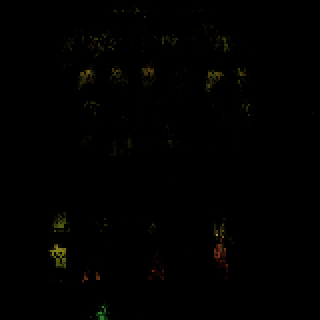} %
    \caption{Three examples of counterfactual states generated using the ablated model.}
    
    \label{fig:ablated_examples}%
\end{figure}

We then created a baseline method consisting of counterfactual images from an ablated version of our generative model. In the ablated version of the network, the encoder, discriminator, and Wasserstein autoencoder were removed, and the generator was trained with MSE loss to reconstruct $\bm{s}$ given $\bm{z}$ as input. Counterfactual images were generated by performing gradient descent with respect to $\bm{z}$ to maximize $\bm{\pi}(\bm{z},a')$ for a counterfactual action $a'$.
We found that counterfactual states generated in this way did not always construct a perfectly convincing game state as shown in Figure \ref{fig:ablated_examples}, but were of sufficient quality to use as a baseline in our user study.
\ref{ablation_descriptions} details other ablation experiments, which reveal the negative effects of removing any specific component from our architecture.

Finally, we also included images from the game itself. In summary, the images in our first user study were generated by three different sources: 10 from the actual game, 10 from our counterfactual state explanation method, and 10 from our ablated network. These images were randomly sorted for each user. 

We evaluated our counterfactual state explanations through a user study in our lab with 30 participants (20 male, 10 female) who were not experts in machine learning; participants included undergraduates and members of the local community. Approximately half were undergraduates and the others were from the community. 80\% were between the ages of 18-30, 10\% were between 30-50, and the other 10\% were between 50-60.  We chose to focus our study on Space Invaders because it is straightforward to learn for a participant unfamiliar with video games. To familiarize participants with Space Invaders, we started the study by having participants play the game for 5 minutes. Participants then rated the fidelity of 30 randomly ordered game images on a Likert scale from 1 to 6: (1) Completely Fake, (2) Most parts fake, (3) More than half fake, (4) More than half real, (5) Most parts real and (6) Completely real.



\subsection{User Study 2: Using Counterfactuals to Detect a Flawed Agent (RQs 2 and 3)} 

Our second user study was intended to evaluate the effectiveness of our counterfactual state explanations. Our focus was on a real world setting in which a user, who was not a machine learning expert, needed to assess a RL agent that was about to be deployed. We designed an objective task that relied on the user's understanding of the agent's decision making process from the counterfactual explanations. The task required participants to identify which of two RL agents was flawed based on the counterfactual explanations provided. As in the first user study, we chose Space Invaders since it was quick to learn and the optimal strategy was not immediately obvious. Since we recruited non-experts in AI or machine learning, we henceforth referred to the RL agent as an AI agent in our user study for simplicity.

The counterfactual explanation's effectiveness was measured by a (2x2)x2 mixed factorial subjects design as we have both a within-subjects comparison and a between-subjects comparison. The within-subjects comparison involved the two independent variables of RL agent type (flawed versus normal) and explanation presence (with and without explanation). Thus, all participants were shown the behavior of the flawed and the normal agents both with and without counterfactual explanations. The between-subjects comparison involved comparing counterfactual explanation methods; one group of participants was shown a baseline counterfactual explanation method based on nearest neighbors and the other group was shown our counterfactual state explanation method.




\subsubsection{Experimental Design}
The participants were presented with the task of identifying which of the two agents was flawed. We designed our two agents such that their average score on the game was almost equal, and the score could not be used to determine which agent was flawed. In addition, humans could not identify the flawed agent by simply watching the agents play the game. Consequently, the counterfactual explanations were the main source of insight into the agent's decision-making for the participants.

An alternative approach for evaluating the effectiveness of counterfactual explanations was to have participants predict an agent's action in a new state. While action prediction may be feasible in some environments (e.g. \cite{madumal2020explainable}), it can also be challenging in other environments such as Atari games and real-time strategy games. \citet{anderson2019mortals} showed that using explanations to predict future actions was difficult, sometimes even worse than random guessing, because AI agents could be successful in these games in ways that were unintuitive to humans.

Our "normal" agent was the agent described in section  \ref{sec:experimental_setup}. For the flawed agent, we tried designing flawed agents that were blind to different parts of the game, but many of these possibilities were easy to detect by humans. Blocking half of the screen resulted in the agent only playing in the visible half. Removing the barriers had no effect as the agent eventually learned their locations during training.
Removing the bullets caused a noticeable behavior change as the agent hid under the barriers for the majority of the game. Finally, we were unable to train an agent that performed well by removing the enemies from the observations. 

We ultimately settled on a flawed Space Invaders agent by masking the region of the screen containing the green ship, effectively making the agent unaware of its own ship's position.  This flaw was subtle and difficult to detect without the aid of counterfactual explanations.

This flawed agent was harder to train than a normal agent playing Space Invaders and thus required 160 million game steps to achieve sufficiently good performance. In addition, for our flawed agent, we set the adversarial loss hyperparameter $\lambda=100$ to make the generated counterfactual states have visually evident changes from the original query state.

\subsubsection{Conditions}

This study involved two conditions corresponding to different counterfactual explanation methods. The first condition used a naive baseline based on a simple nearest neighbor approach. The second condition involved our counterfactual state explanations.

\paragraph{{\bf Nearest Neighbor Counterfactual Explanations (NNCE)}} For this approach, the agent played the game for $N=25$ million time steps with $\epsilon$-greedy exploration to produce a game trace dataset $\mathcal{\bm{D}}$, which we used for nearest neighbor selection. For each step we stored in $\mathcal{\bm{D}}$, the state $\bs$, the representation $\bz = A(\bs)$, and the action taken $a$, resulting in a dataset $\mathcal{D} = \{(\bm{s}_1, \bz_1, a_1), \ldots, $
$(\bm{s}_N, \bz_N, a_N))$. 
To generate a counterfactual from this dataset, the agent played a new game and on the desired query state $\bs$ we found the nearest latent point $\bz^* \in \mathcal{D}$ to the current point $\bz = A(\bs)$ where the agent took the desired action of $a'$; we used $L_2$ distance to determine closeness. We then displayed the associated state $\bs^*$ from the triplet $(\bs^*,\bz^*,a')$ as the closest counterfactual state where the agent took a different action $a'$. Note that the images from the nearest neighbor approach were always faithful to the game as they were actual game frames from the Atari game. However, even with a very large game trace dataset of size 25 million, the nearest neighbor approach did not always retrieve a game state that was "close" to the query state. In contrast, our counterfactual state explanations were always close to the query state by design, but they may not always have complete fidelity to the game.

\paragraph{\bf Choosing counterfactual query states and counterfactual actions} The specific images, serving as query states to present to participants for our counterfactual state explanations, were objectively chosen using a heuristic based on the entropy of the policy vector $\bm{\pi}(A(s))$ of state $s$; this entropy score has been used in the past for choosing key frames for establishing trust \citep{huang2018}. For diversity, if an image at time $t$ was selected, we do not allow images to be selected until after time $t+10$. This restriction was especially important for diversity in the counterfactual states chosen for the flawed agent as it had very low entropy in its policy vector from the initial states, but higher entropy later on. 
As Space Invaders is a relatively simple game in which the aliens move faster as time progresses, we only considered diversity in terms of the progression of time within a round and we selected query states at different points in time. All query states used in the study can be seen in Appendix figures \ref{fig:space_allimgs_normal1} - 
\ref{fig:space_allimgs_abl2}.
We thus emphasize the fact that the counterfactual states and corresponding actions we presented to participants were not hand-picked; rather, they were selected objectively by our heuristic. 

For our counterfactual state explanations, once a query state was selected, we chose the counterfactual action $a'$ as the one that involved the largest $L_2$ change between the original Wasserstein latent state $\bm{z_w}$ and the counterfactual Wasserstein latent state $\bm{{z_w}'}$ (ignoring the no-operation action). 

For NNCE in our user study, we use the same entropy-based state selection heuristic to determine which query states to show to the participant, thereby ensuring that query states are identical between the two conditions. What varies between the two conditions is the explanation process, which selects the counterfactual action $a'$ and the resulting counterfactual state $\bm{s'}$. The method we used for selecting the counterfactual action $a'$ for NNCE differs from the heuristic used by our counterfactual state explanations. To select the counterfactual action in NNCE, we find the closest nearest neighbor in latent space $\bz$ (via $L_2$ distance) where the agent performs a different action $a' \neq a$. 

The action selection heuristics were slightly different between the two conditions in order to maximize the quality of selected counterfactual states by the different methods. The two methods differed in how far the counterfactual images were from the query state due to different latent spaces being used by the two methods and also due to the granularity of their movement in their respective latent spaces. Our counterfactual state explanations operated in a Wasserstein latent space. Due to the fact that they were created by a generative process and not retrieved from a dataset, using the closest Wasserstein point with a different action often caused very little change or no change at all. In contrast, the NNCE method operated in the latent space of the pre-trained agent, which does not have a Wasserstein latent space. The NNCEs used pre-existing images from the dataset $\mathcal{\bm{D}}$, which were usually further away from the query state (visually) than most counterfactual states created by our method. Had we chosen the counterfactual action that involved the largest $L_2$ change in latent space, NNCEs would have produced images that were often dramatically different from the query state, which would likely have produced worse results in our user study. Instead, to give the NNCE condition the best possible counterfactual images (based on visual inspection), we ultimately selected the counterfactual action to be the action (different from the original action $a$) associated with the nearest neighbor with the closest $L_2$ distance in latent space.



\subsubsection{Participants and Procedure}

We recruited 60 participants at Oregon State University, with 30 participants in each condition. The target audience for our user study was people who were not experts in machine learning. Approximately half were undergraduates and the others were from the community. All participants were between the ages of 18-40, 40\% of whom were women and 60\% were men. 
This study consisted of 6 sections:
\begin{enumerate}
	\item Gameplay
	\item Agents Analysis (pre-evaluation)
	\item Tutorial
	\item Evaluation (main task)
	\item Agents Analysis (post-evaluation)
	\item Reflection
\end{enumerate}



\paragraph{1. Gameplay} 
A facilitator started the study with a guided tutorial about game rules and described the task to be performed, after which the participants were allowed to use the system. To be able to understand the game better, all the participants first played the Atari 2600 video game Space Invaders for 5 minutes.

\paragraph{2. Agents Analysis (pre-evaluation)} 
After having enough hands-on experience with the game, each participant watched a video of the normal agent and a video of the flawed agent playing one complete episode of the game from start to finish. The identities of each agent were hidden from participants. The videos were selected such that the agent cleared all enemies before they reached the bottom while avoiding all incoming bullets. We randomized the order of presentation of the normal and flawed agent. For concreteness, we described the flawed agent to the participants as an agent with a malfunction in its sensors.  After viewing the videos, we  asked the participants, ``Which of the two AI do you believe has a malfunction?", with their choices being "AI one", "AI two", or "CAN'T TELL". We then asked the participants if they could identify which part of the game was the flawed AI blind to: the yellow aliens, the white bullets, the green ship, or the orange barriers.  After answering both questions, participants were placed on a waiting screen to ensure the next section occurred simultaneously for everyone. At this point, participants were unable to change their answers to the previous questions and were unable to view the videos for the rest of the study. 

The answers from this section formed the data corpus of the participants' descriptive analysis of the AI agents \textbf{before} they saw the explanation. 

\begin{figure}[t]
    \centering
    \includegraphics[width=.99\linewidth]{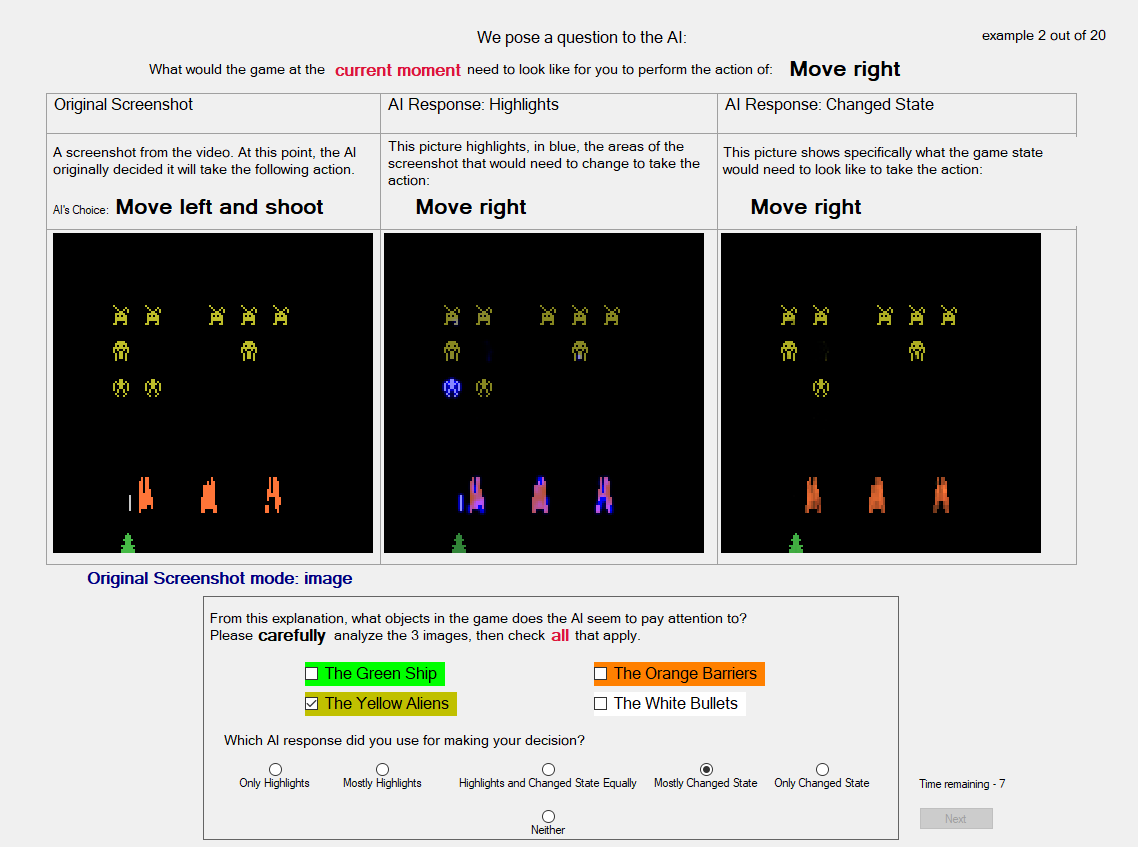}
   
   \caption{The explanation tool used for displaying counterfactual states to participants in our user study. }
    
    \label{fig:ex_tool}%
\end{figure}
\paragraph{3. Tutorial}
The facilitator then gave a detailed guided tutorial to describe the counterfactual explanation display tool, particularly since counterfactuals are an esoteric topic for most non-experts. 
The display tool consisted of a tuple of 3 images: the original state where an agent took its preferred action $a$, the changed state where the agent took the action $a'$, and the image highlights. The original state was an interactive UI element, where if clicked it would change into a GIF that sequentially displayed the current and three previous game steps to give context for the query state. We omitted the term "counterfactual" as we found the additional vocabulary to be confusing in our preliminary studies. 

Our approach of presenting the original image, highlights and the counterfactual alongside each other ties in with the causal connection of abductive reasoning and inference \citep{TimMillerSocialSciencePaper2019}, where the highlights served the purpose of ``fixing the eyeballs'' of the participants, and the counterfactuals offered reasoning for the AI's actions.
The exact script for the tutorial is provided in the appendix section \ref{sec:script}.

\paragraph{4. Evaluation (main task)}
Following the tutorial, the participants evaluated 20 screenshots of the game in the display tool: 10 states for each agent selected via the heuristic described above. All 10 states were selected from the single game episode shown in the videos from earlier. An example of a screen in the explanation tool can be seen in figure \ref{fig:ex_tool}.

For each screenshot, we asked the participant two questions. The first question being "What objects do you think this AI pays attention to?" with 4 check boxes to be potentially selected (aliens, bullets, ship, barriers) and the second question being "Which explanation did you use for making your decision?" with 6 options (Only Highlights, Mostly Highlights, Highlights and Changed State Equally, Mostly Changed State, Only Changed State, or Neither). We presented the 20 explanations (of normal and flawed agents) in randomized order to avoid biasing the participants. 

The answers from this section formed the data corpus of the participants' descriptive analysis of the AI agents \textbf{after} they saw the explanation. 



\begin{figure}[t]
    \centering
    \includegraphics[width=.85\linewidth]{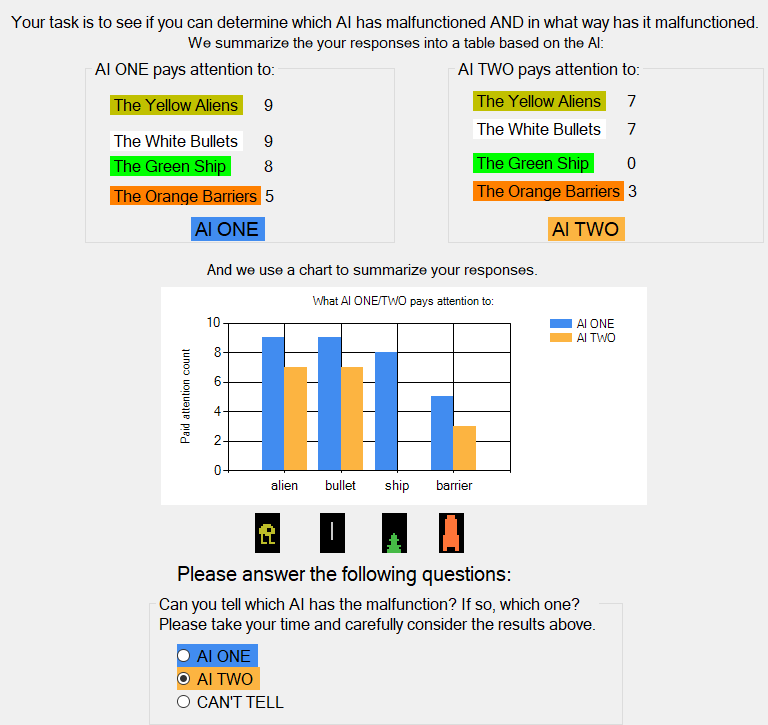}
   
   \caption{An example of the results screen a user would see after completing the evaluation for the Counterfactual States condition.}
    
    \label{fig:results_screen}%
\end{figure}
%


\paragraph{5. Agents Analysis (post-evaluation)}
Once the participants finished evaluating the 20 explanations, we summarized the results of their own responses to the question "What objects do you think this AI pays attention to?" in a table and a chart, separating the two different agents and tallying the number of times the participant selected each object. We then re-asked the same questions from study section two: which AI is malfunctioning and in what way is it malfunctioning.
An example of the final results screen can be seen in Figure \ref{fig:results_screen}, where each vertical element of the UI was hidden until the user clicks a ``continue'' button to guide participants through the summary data one step at a time. 
We found that only showing the tallied results before re-asking the questions to be the best way to get participants to focus on the explanations. In our preliminary experiments leading to the design of the final study, we found that participants were overwhelmed with data if they could go back to either look at the individual examples or re-watch the videos of the agents playing the game.

\paragraph{6. Reflection}
We ended the study by asking the participants to perform a short written reflection after they submitted their answer to gauge their understanding of the explanations, and to elicit their opinion on the explanation. The questions included, ``Which parts of the explanation tool influenced your decision in determining the malfunctioning AI?'' to understand what participants found helpful in the explanation, and what contributed to successfully finding the flawed agent. We also asked the participants to describe components of the explanation, ``In your own words, can you briefly explain what the 3$^{rd}$ image from the explanation tool is (the images titled: "AI Response Changed State")?'' to gauge if participants even understood the concept of a counterfactual reasonably well, and how they had done in the main task if they did not. \ref{sec:coding} describes the content analysis applied to these two questions.

%% file: results_imgs.tex
\subsection{Example Counterfactual States}

We now show examples of counterfactual states for pre-trained agents in various Atari games; these examples include both high and low quality counterfactuals. In Figures \ref{fig:qbert_imgs} to \ref{fig:space_generative_normal},
we show sets of images in which the left image is the original query state where the agent would take action $a$ according to its policy, the right image is the counterfactual state where the agent would take the selected action $a'$, and the center image is the highlighted difference between the two.

\subsubsection{$\text{Q}^*$bert}

\begin{figure*}[h]
    \centering
    
    \subfloat{
    \includegraphics[width=.8\linewidth]{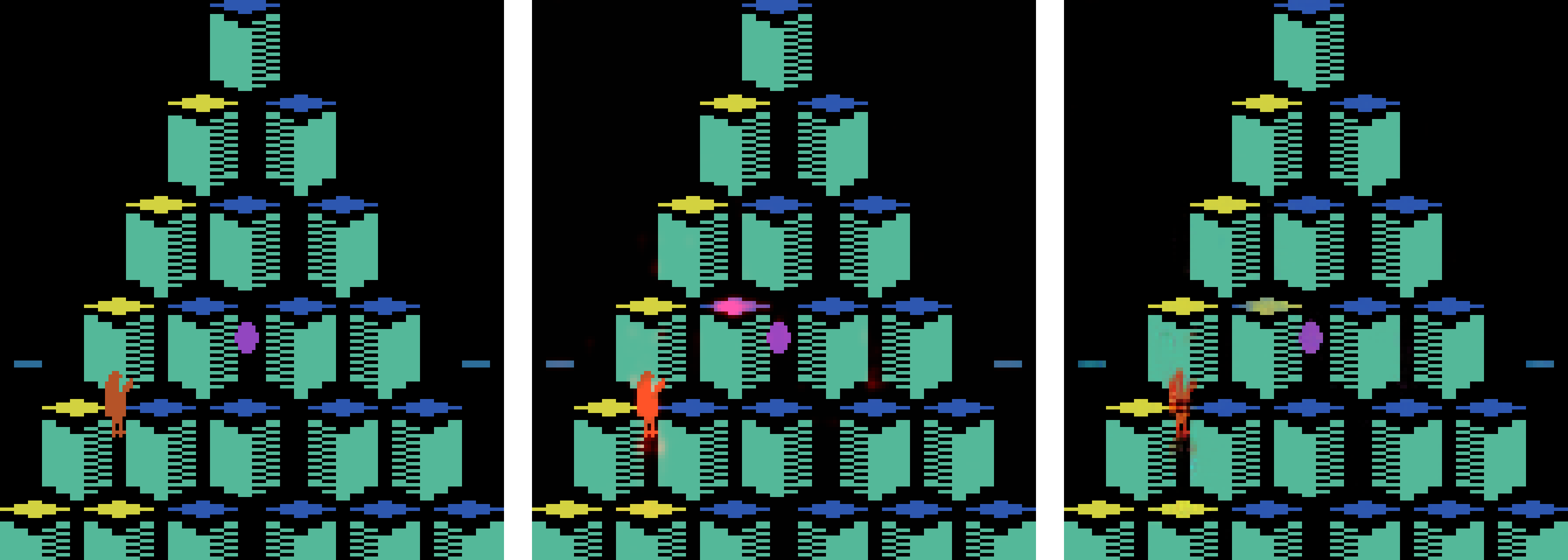} 
    
    }
    \vspace{-2ex}
    \caption*{$a=\text{MoveUpRight}$, $a'=\text{MoveUpLeft}$}
    \vspace{-2ex}
    \subfloat{
    \includegraphics[width=.8\linewidth]{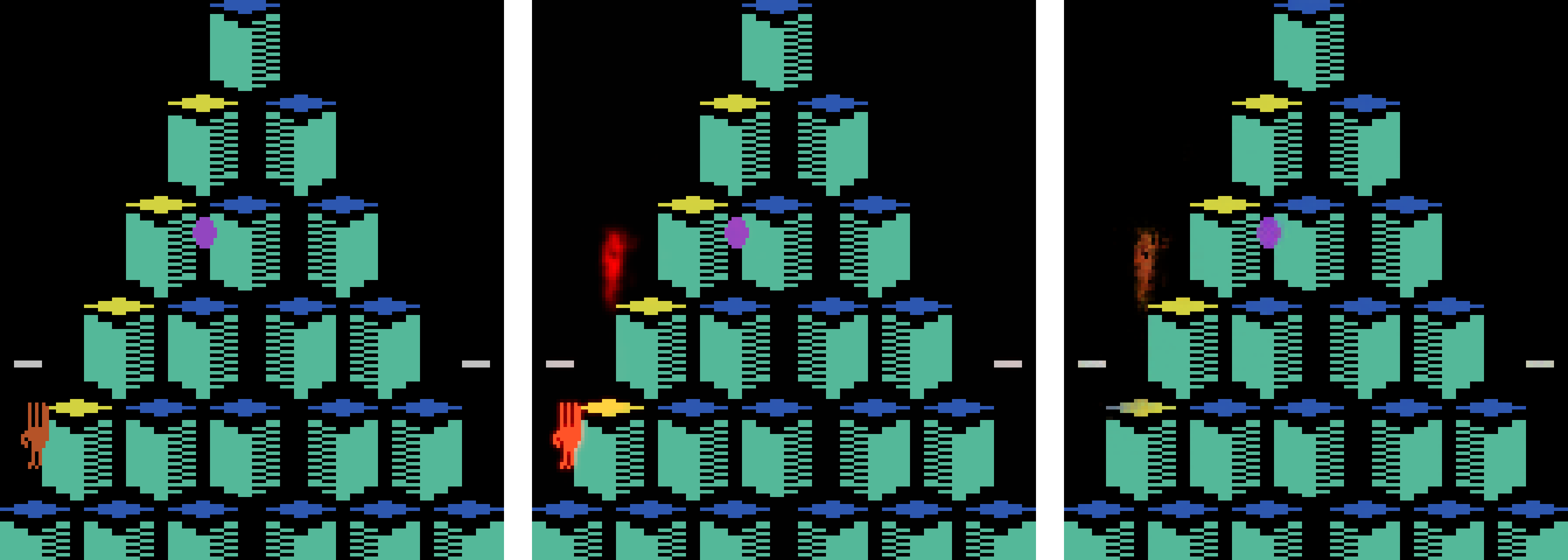} 
    }
    \vspace{-2ex}
    \caption*{$a=\text{MoveUpRight}$, $a'=\text{MoveDownLeft}$}
    \vspace{-2ex}
    \caption{
    Each row shows an example of a counterfactual state explanation for \textit{$\text{Q}^*$bert}: Query state with action $a$ (\textbf{left}), counterfactual state with action $a'$ (\textbf{right}), and red highlights (\textbf{center}).
     }

    \label{fig:qbert_imgs}%
\end{figure*}
In this game, the agent controls the orange character Q*bert, who starts each game with 3 lives at the top of a pyramid and has 5 actions to hopping diagonally from cube to cube (or stay still). Landing on a cube causes it to change color, and changing every cube to the target color allows the agent to progress to the next stage. The agent must avoid purple enemies or lose a life upon contact. Green enemies that revert cube color changes can be stopped via contact.

In the top row of Figure \ref{fig:qbert_imgs}, the counterfactual shows that if the up-right square were yellow (already visited), Qbert would move up-left. In the bottom row of Figure \ref{fig:qbert_imgs}, if Qbert had been higher up on the structure, the agent will jump down and left; in this example, the Qbert image is not perfectly realistic but enough to give a sense of the agent's decision making.

\subsubsection{Seaquest}
\begin{figure}[t]
    \centering
    \subfloat{\includegraphics[width=.8\linewidth]{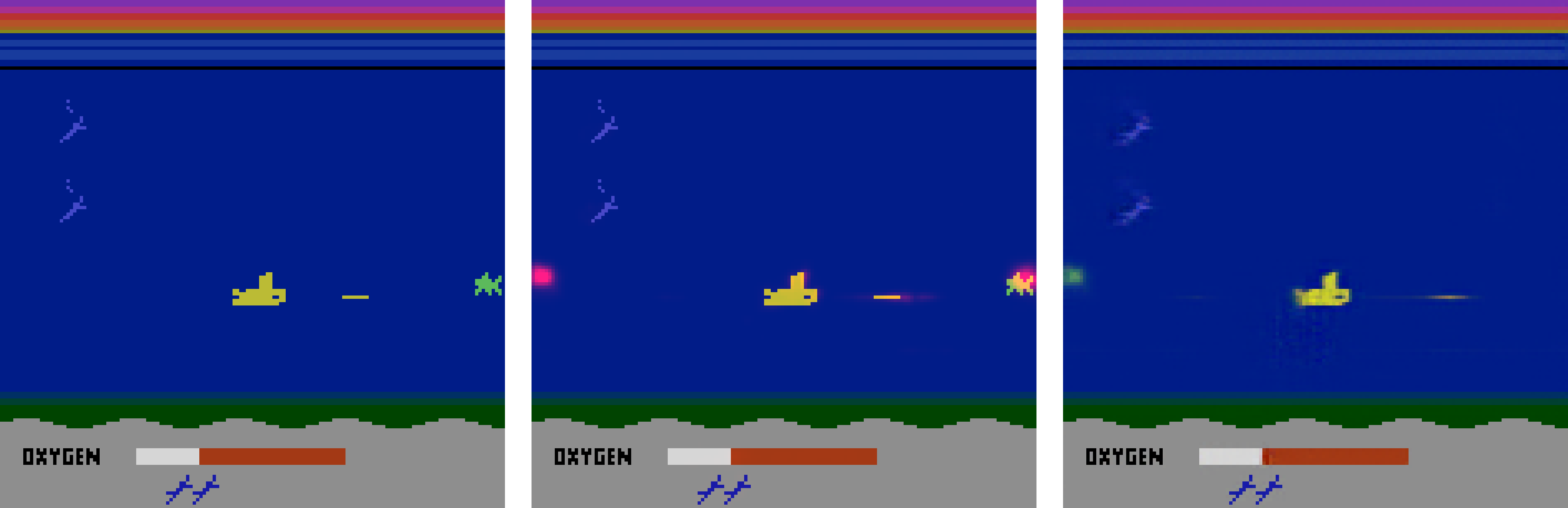} }
    \vspace{-2.5ex}
    \caption*{$a=\text{MoveUpRightAndShoot}$, $a'=\text{MoveUpLeftAndShoot}$}
    \vspace{-2.5ex}
    \subfloat{\includegraphics[width=.8\linewidth]{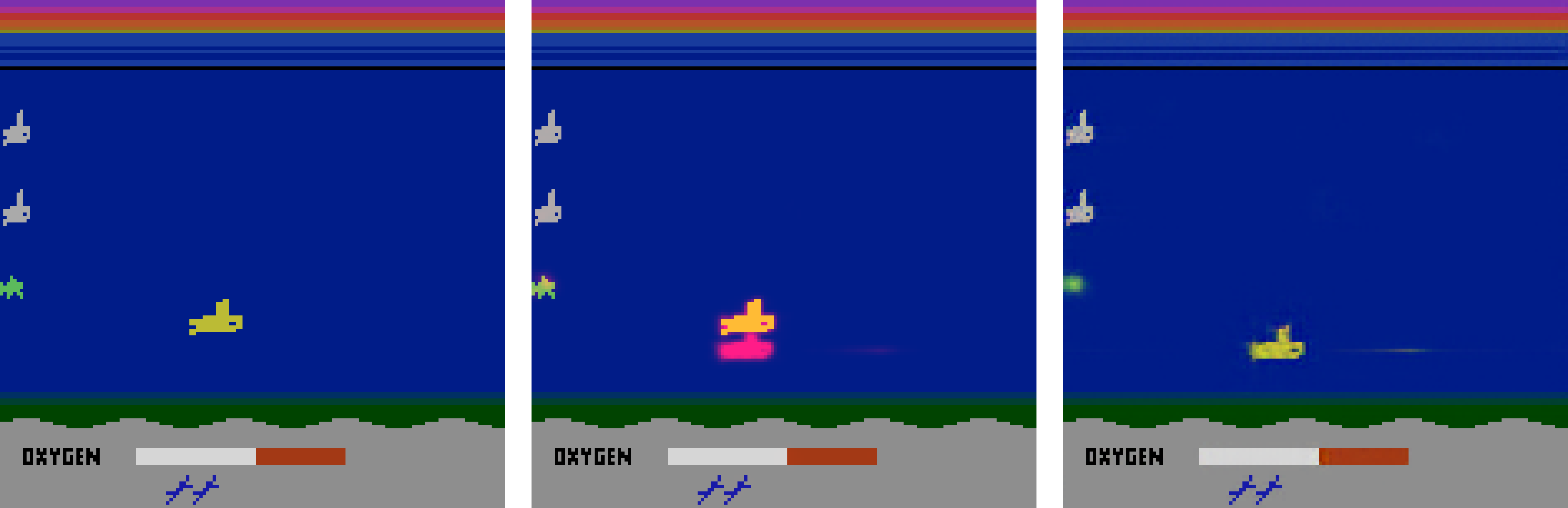} }
    \vspace{-2.5ex}
    \caption*{$a=\text{MoveUpLeftAndShoot}$, $a'=\text{MoveUpLeft}$}
    \vspace{-2.5ex}
    \subfloat{\includegraphics[width=.8\linewidth]{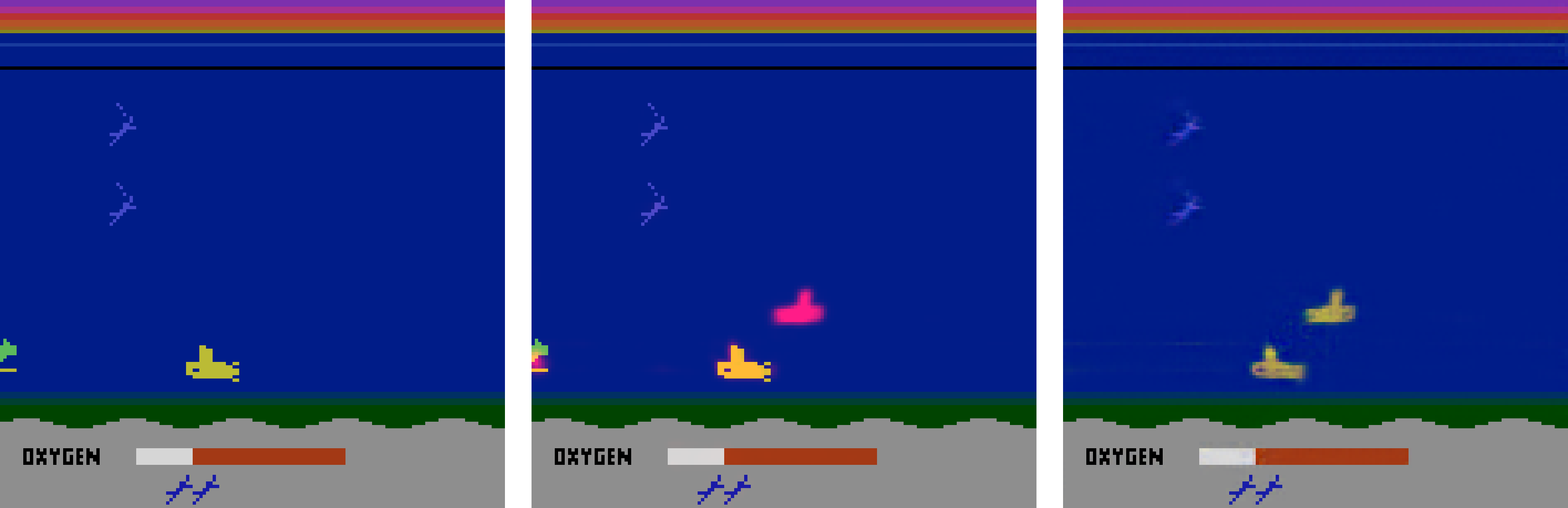} }
    \vspace{-2.5ex}
    \caption*{$a=\text{MoveLeftAndShoot}$, $a'=\text{MoveDown}$}
    \vspace{-2.5ex}
\caption{Each row shows an example counterfactual state explanation for \textit{Seaquest}: Query state with action $a$ (\textbf{left}), counterfactual state with action $a'$ (\textbf{right}), and highlights (\textbf{center}).
     }

    \label{fig:seaquest_imgs}%
\end{figure}
In this game, an agent must shoot torpedoes at oncoming enemies while rescuing friendly divers. 
In Figure \ref{fig:seaquest_imgs} (top row), a new enemy must appear to the left in order for the agent to take an action that turns the submarine around while firing. Thus, the agent has an understanding about enemy spawns and submarine direction. The middle row of Figure \ref{fig:seaquest_imgs} shows a scenario (best viewed on a computer) where the agent would move up and left but not shoot because the agent would not be fully aligned with the enemy fish on the left to hit it; in addition, the submarine has already shot its torpedo in anticipation of an enemy fish appearing on the bottom right and there can only be one torpedo on the screen at a time. Note that the torpedo is actually highlighted in red but due to the size of the image in Figure \ref{fig:seaquest_imgs}, these highlights are imperceptible.


Figure \ref{fig:seaquest_imgs} (bottom row) shows an unrealistic counterfactual, where despite never seeing two submarines in the training data, the best prediction of the Generator (given the counterfactual inputs), is to place a submarine at both locations.
 

\subsubsection{Crazy Climber}
\begin{figure}[htp]
    \centering
    \subfloat{\includegraphics[width=.8\linewidth]{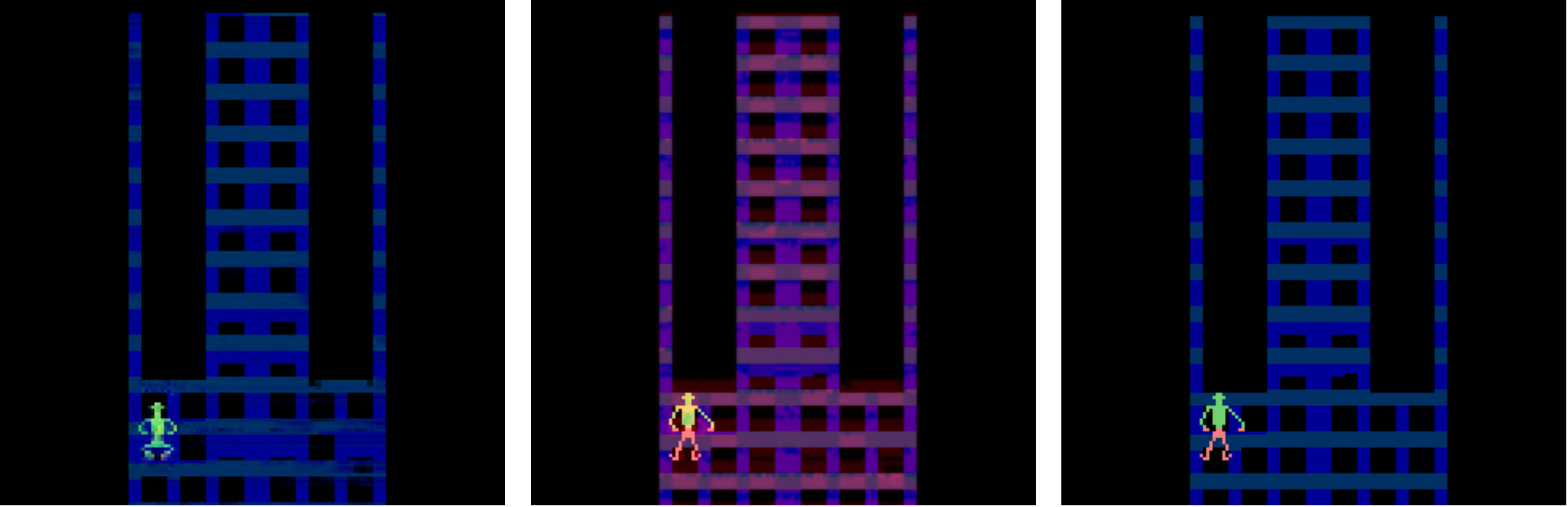} }
    \vspace{-2ex}
    \caption*{$a=\text{MoveRight}$, $a'=\text{MoveBodyUp}$}
    \vspace{-2ex}
    \subfloat{ \includegraphics[width=.8\linewidth]{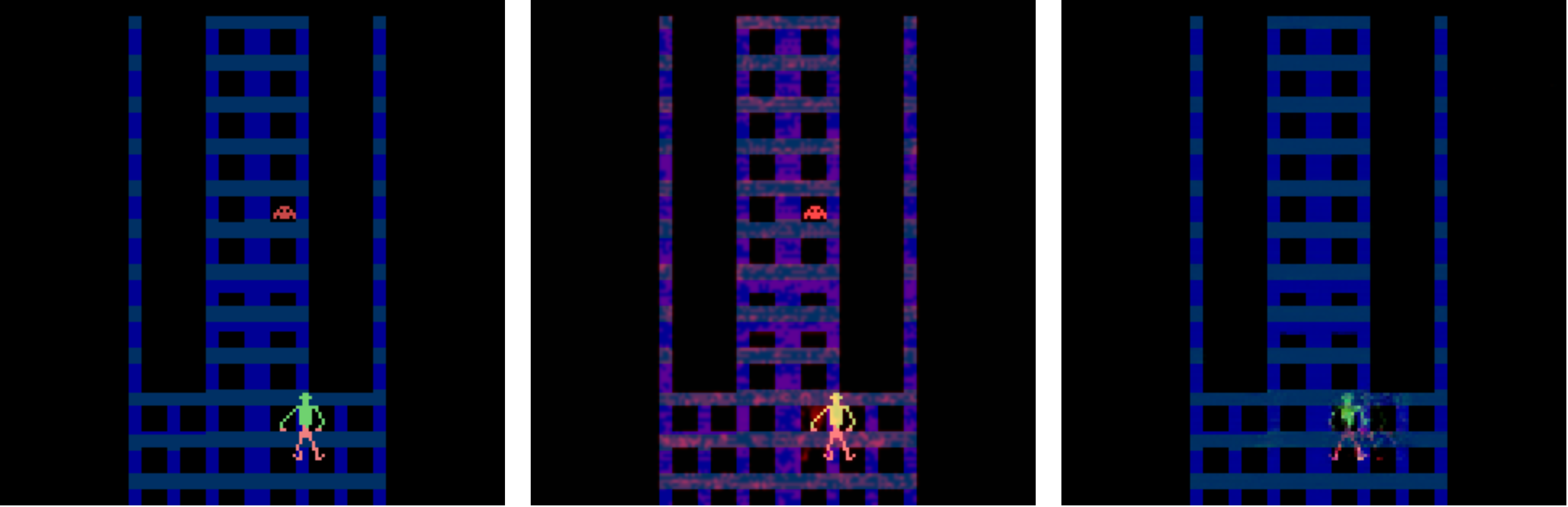} }
    \vspace{-2ex}
    \caption*{$a=\text{MoveLeft}$, $a'=\text{MoveArmsUp}$}
    \vspace{-2ex}
    \caption{
    Each row shows an example counterfactual state explanation for \textit{Crazy Climber}: Query state with action $a$ (\textbf{left}), counterfactual state with action $a'$ (\textbf{right}), and highlights (\textbf{center}).
     }

    \label{fig:cc_imgs}%
\end{figure}
In this game, an agent must climb up a building while avoiding various obstacles. Figure \ref{fig:cc_imgs} (top row) shows the original state in which the agent is in a position to move horizontally, whereas the counterfactual state shows the climber in a ready state to move vertically as indicated by the position of its legs. Figure \ref{fig:cc_imgs} (bottom row) demonstrates how the agent will climb up as the enemy is no longer above it. 
For both examples, because the climber stays in a fixed vertical position with the entire tower itself moving down, the highlights are difficult to interpret. These examples show the importance of using both the highlights and the counterfactual states as in some cases, the counterfactual states are much easier to understand than the  highlights. 

\subsubsection{Space Invaders}

\begin{figure}[htb]
    \centering
    \includegraphics[width=.3\linewidth]{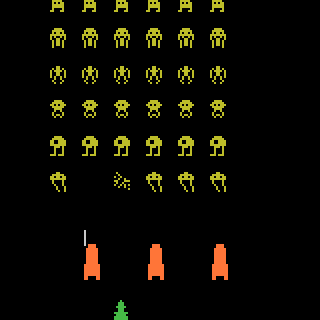} 
    \quad
    \includegraphics[width=.3\linewidth]{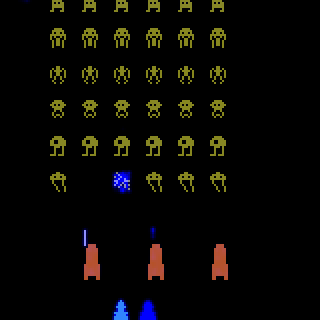} 
    \quad
    \includegraphics[width=.3\linewidth]{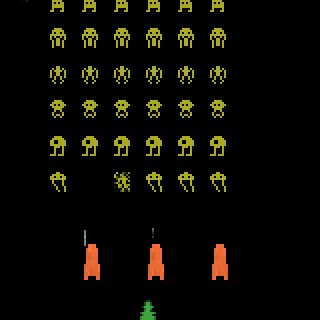} 
   
    \caption{
     An example of a counterfactual state explanation for Space Invaders with the "normal" agent. Here, action $a=\text{MoveRightAndShoot}$ (\textbf{left}), counterfactual state where action $a'=\text{MoveRight}$ (\textbf{right}), and the highlighted difference (\textbf{center}).
         }
    \label{fig:space_generative_normal}%
\end{figure}

In this game, an agent exchanges fire with approaching enemies while taking cover underneath three barriers. Figure \ref{fig:space_generative_normal} depicts the example, which was also used in our user study. This example reveals that the agent has learned to prefer specific locations for safely lining up shots, selectively choosing enemies to shoot.

\begin{figure}[t]
    \centering
    \includegraphics[width=.3\linewidth]{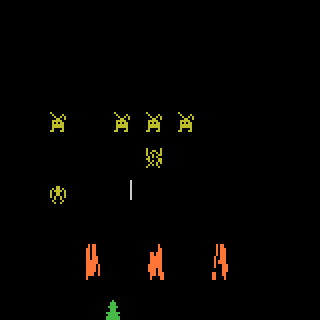}
   \quad
    \includegraphics[width=.3\linewidth]{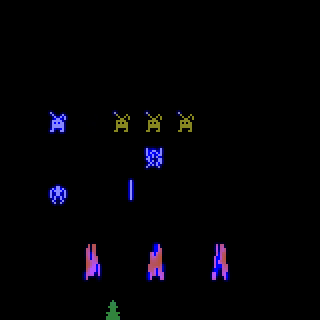}
   \quad
    \includegraphics[width=.3\linewidth]{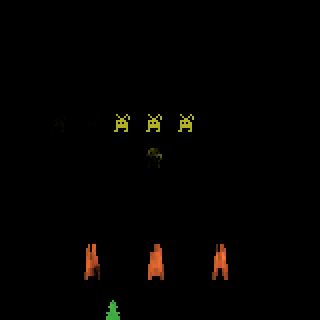}
    \caption{
    An example of a counterfactual state explanation for Space Invaders with the flawed agent from our second user study. Here, action $a=\text{MoveLeftAndShoot}$ (\textbf{left}), counterfactual state where action $a'=\text{MoveRight}$ (\textbf{right}), and the highlighted difference (\textbf{center}).
         }

    \label{fig:space_generative_malf}%
\end{figure}

We also include an example of a counterfactual state explanation with the flawed agent in our second user study. Figure \ref{fig:space_generative_malf} shows that in the generated counterfactual state explanation, the flawed agent does not move the ship as it is blind to its own ship's location; in fact, the flawed agent never moves the ship in all of our counterfactual state explanations. 

%% file: results_user_study.tex
\subsection{User Study Results}

\subsubsection{RQ 1: Fidelity of Counterfactuals}
\label{sec:realism}

\begin{table}[htb]
\centering
\begin{tabular}{ |c|c|c|c| } 
\hline
 & {\bf Ablated} & {\bf Counterfactual State} & {\bf Actual Game} \\
 & {\bf Version} & {\bf Explanations} & {} \\
\hline
{\bf Score} & 1.93 & 4.00 & 4.97\\
\hline
\end{tabular}
\caption{Average results on a 6 point Likert scale from the fidelity user study.}
\label{table:realism}
\end{table}

In terms of fidelity, the average ratings on the 6 point Likert scale are shown in Table \ref{table:realism}. The differences between the fidelity ratings for the counterfactual states and real states were not statistically significant ($\alpha=0.05$, p-value=0.458, one-sided Wilcoxon signed-rank test). These results show that our counterfactual states were on average close to appearing faithful to the game states but they were not perfect. In the next section, we will show that despite these imperfections, the counterfactuals were still useful to participants.



\subsubsection{RQ 2: Can counterfactual states help users identify a flawed agent?}


Participants were significantly more successful at identifying the flawed agent when provided with counterfactual explanations for both the counterfactual state explanations ($\alpha$ = 0.05, p-value = 0.0011, Pearson's Chi-square test) and the NNCEs ($\alpha$ = 0.05, p-value=0.0009, Pearson's Chi-square test).

This hypothesis was further reinforced when all participants in both conditions self-reported to have found the explanation useful in the evaluation section. Only 1 participant out of 60 stated that the video in the Agents Analysis section was useful. Instead, participants found the highlights and the counterfactuals to be more useful than the video.





\begin{figure}[htb]
\begin{center}
\includegraphics[width=0.99\columnwidth]{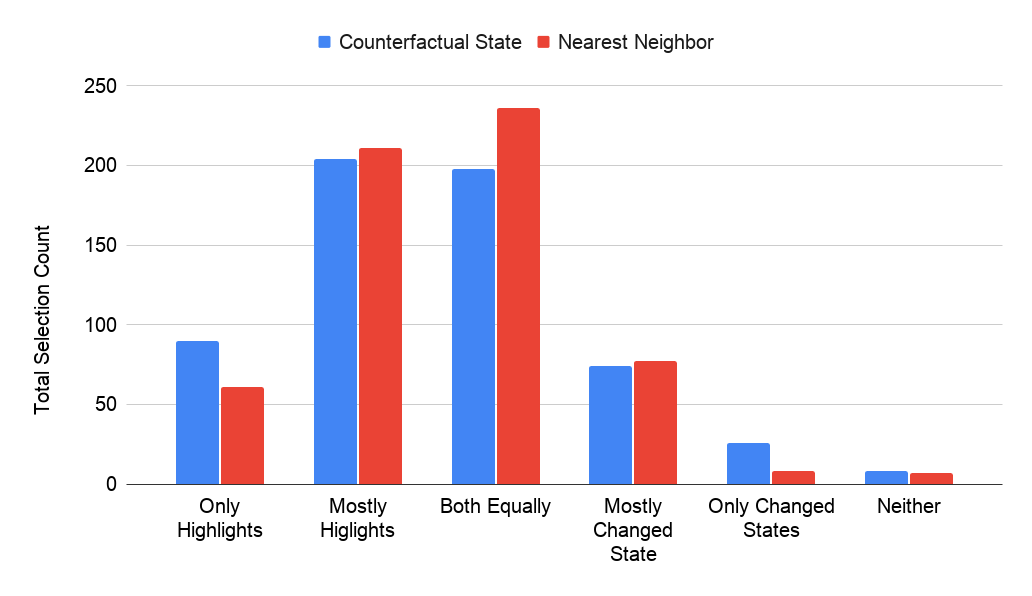}
\caption{The total selection count over all participants and all explanations regarding the self-reported usefulness of each explanation component.}
\label{fig:response_histogram}
\end{center}
\end{figure}

In the evaluation section, for each explanation from a given counterfactual method, we asked participants to rate the usefulness of each component of the explanation on a 5 point Likert scale (1: Highlights only, 2: Mostly Highlights, 3: Both Equally, 4: Mostly Counterfactuals, 5: Counterfactuals only). For counterfactual state explanations, ``Mostly Highlights'' was the most common response (204/600 times; 34\%) for helping participants identify the flaw in the AI. For NNCE, ``Both Equally'' was the most common response (236/600 times; 39\%). The full response distribution for each condition is shown in Figure \ref{fig:response_histogram}. These results indicate that neither component in isolation was ideal. Most of the time, participants preferred having both, but with varying degrees of usefulness. We also found this result in the qualitative data from the post task questionnaire, wherein participants in both conditions overwhelmingly self-reported to have used the highlights as a supporting artefact for the counterfactual explanations, and vice-versa:

\begin{displayquote}\textbf{Participant 43 in Counterfactual State condition: }
``I used the highlights tool primarily because it was the easiest way to see what was changing from the original state. Then, I would reference the changed state tool to see how the original changed.''
\end{displayquote}

\begin{displayquote}\textbf{Participant 14 in Nearest Neighbor condition: }
``Mostly highlights, I used changed state sparingly to cement assertions from the highlights.''
\end{displayquote}




Participants in both conditions found the summary chart to be helpful in consolidating their ideas and facilitating recall. For instance, two participants commented in their responses:

\begin{displayquote}\textbf{Participant 35 in Counterfactual State condition: }
``The bar graph at the end of my responses for both AIs influenced it the most.''
\end{displayquote}

\begin{displayquote}\textbf{Participant 16 in Nearest Neighbor condition: }
``The charts at the end heavily influenced my decision, because I thought the malfunctioning AI couldn't see the barriers because they had more damage on the ship side of the barriers than the alien side, but the charts showed that that was a poor assumption because almost every time I evaluated the barriers as something they could see.''
\end{displayquote}





\subsubsection{RQ 3: Comparison of Counterfactual methods}

\begin{table}[htb]
\centering
\begin{tabular}{|c|c|c|c|}
\hline
{} & {\bf Incorrect} & {\bf Correct} & {\bf Can't tell} \\
{} & {\bf Identification} & {\bf Identification} & {} \\
\hline
{\bf Without explanation }  & 10 ($33\%$)                                                                & 17 ($57\%$)                                                                  & 3 ($10\%$)         \\ \hline
{\bf With explanation}       & 2 ($7\%$)                                                                 & 27 ($90\%$)                                                                  & 1 ($3\%$)         \\ \hline
\end{tabular}
\caption{The number of participants, with and without counterfactual state explanations, who incorrectly identified the normal AI, correctly identified the flawed AI and who were unable to tell the difference.}
\label{table:cs}%
\end{table}

Participants that were provided with counterfactual state explanations identified the flawed AI with a far higher success rate than the NNCEs. Without any explanations, $57\%$ of the participants correctly identified the flawed agent (Table \ref{table:cs}). With counterfactual state explanations, this percentage improved to $90\%$, which is a significant improvement at the ($\alpha=0.05$, p-value = $10^{-9}$, Pearson's Chi-square test).  In addition, none of these participants were able to correctly determine the specific flaw in the agent in the first Agent Analysis section. However, after using our counterfactual state explanations, $60\%$ of participants accurately diagnosed the specific flaw, which is statistically significant ($\alpha=0.05$, p-value $=$ 0, Pearson's Chi-square test).




\begin{table}[htb]
\centering
\begin{tabular}{|c|c|c|c|}
\hline
{} & {\bf Incorrect} & {\bf Correct} & {\bf Can't tell} \\
{} & {\bf Identification} & {\bf Identification} & {} \\
\hline
{\bf Without explanation }       & 9 ($30\%$)                                                                 & 19 ($63\%$)                                                                   & 2 ($7\%$)\\ \hline  
{\bf With explanation} & 9 ($30\%$)                                                                 & 14 ($47\%$)                                                                  & 7 ($23\%$)         \\ \hline
\end{tabular}
\caption{The number of participants, with and without NNCEs, who incorrectly identified the normal AI, correctly identified the flawed AI and who were unable to tell the difference.}
\label{table:nn}%
\end{table}

\begin{figure}[t]
\begin{center}
\includegraphics[width=0.85\columnwidth]{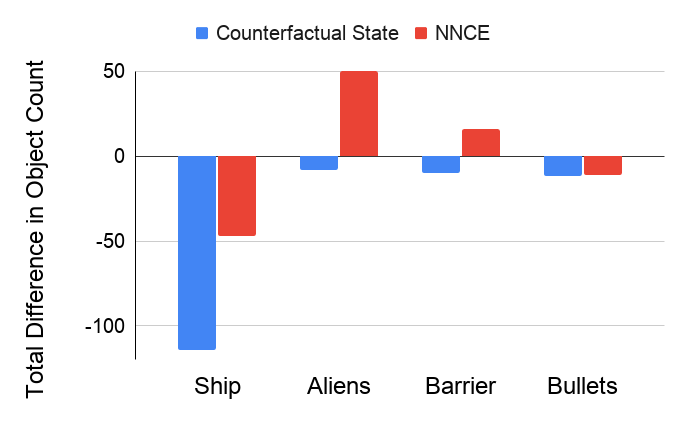}
\caption{The total difference in object counts over all participants, where the \emph{object} refers to the Space Invaders element that a participant determines the agent pays attention to. Here, the y-axis measures the difference between the total object counts for the flawed agent minus the total object counts for the normal agent. Positive numbers indicate that participants consider the flawed agent to pay more attention to that object than the normal agent, whereas negative numbers indicate that participants consider the flawed agent to pay less attention to that object.}
\label{fig:objects_histogram}
\end{center}
\end{figure}


In contrast, NNCEs often confused participants. $63\%$ of the participants identified the flawed agent correctly with just the video (Table \ref{table:nn}), but after viewing the explanations, this percentage dropped to $47\%$ ($\alpha=0.05$, p-value = 0.1432, Pearson's Chi square test). Figure \ref{fig:objects_histogram} contains an aggregate comparison of how well participants that were shown NNCE versus counterfactual state explanations could identify the specific flaw. The histogram in Figure \ref{fig:objects_histogram} depicts the difference in object counts over all participants, where the object refers to the Space Invaders element that a participant determines the agent pays attention to. The difference is computed as the total object counts for the flawed agent minus the total object counts for the normal agent.  Participants for both counterfactual approaches were able to pick up on the correct flaw, but participants that were shown counterfactual state explanations did so with much higher numbers than participants that were shown NNCEs.

\begin{figure}[t]
    \centering
    \includegraphics[width=.3\linewidth]{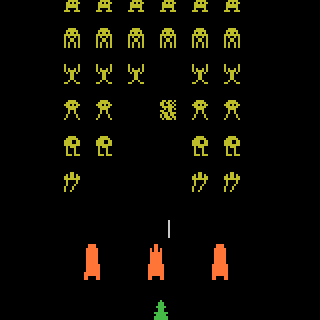} 
    \quad
    \includegraphics[width=.3\linewidth]{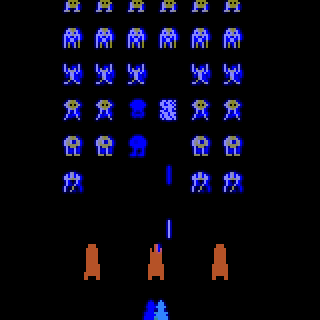}
    \quad
    \includegraphics[width=.3\linewidth]{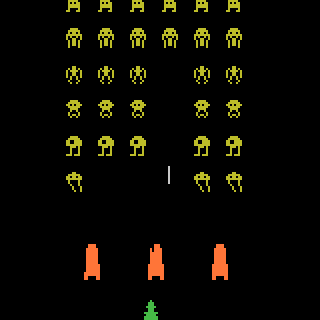}
   
    \caption{
    An example of the nearest neighbor counterfactual explanation method for the normally trained agent with query state where action $a=\text{MoveRightAndShoot}$ (\textbf{left}), counterfactual state where action $a'=\text{MoveLeftAndShoot}$ (\textbf{right}), and the highlighted difference (\textbf{center}).
         }
     \label{fig:space_nn_normal}%
\end{figure}

\begin{figure}[t]
    \centering
    \includegraphics[width=.3\linewidth]{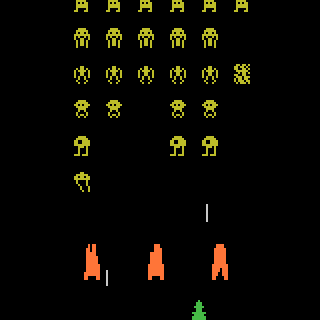}
    \quad
    \includegraphics[width=.3\linewidth]{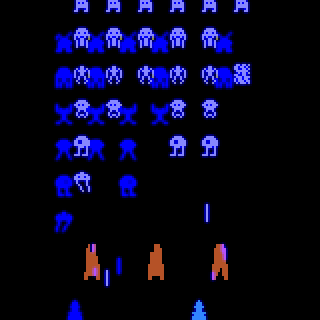}
    \quad
    \includegraphics[width=.3\linewidth]{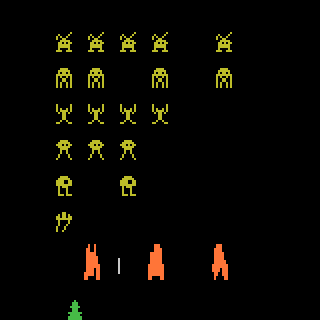}
   
    \caption{
    An example of the nearest neighbor counterfactual explanation method for the flawed agent with query state where action $a=\text{MoveLeftAndShoot}$ (\textbf{left}), counterfactual state where action $a'=\text{MoveLeft}$ (\textbf{right}), and the highlighted difference (\textbf{center}).
         }
    \label{fig:space_nn_malf}%
\end{figure}

One of the main reasons for this decrease is that the NNCEs are inconsistent in quality as the quality depended on the existence of an instance in the game trace dataset $\mathcal{\bm{D}}$ that was reasonably close (in latent space) to the query state. Despite a very large game trace dataset (25 million game frames) as a pool for the NNCEs, a suitable instance that can serve as a counterfactual may not exist, resulting in odd changes to the query state (e.g. an extra alien appearing on the opposite side from the agent) or counterfactuals that were extremely different from the current state (e.g. a reset of the game has occurred or many enemies have been added/removed). Examples of low quality nearest neighbor counterfactuals can be seen in Figures \ref{fig:space_nn_normal} and  \ref{fig:space_nn_malf}. Note that both examples have a large number of highlights. 
In addition, the NNCE in Figure \ref{fig:space_nn_malf} actually moves the ship, which obscures the true flaw in the agent. This reliance on finding a suitable counterfactual in the game trace dataset is a major disadvantage of nearest neighbor counterfactuals. It is likely infeasible to generate a large enough dataset to facilitate retrieval of a reasonable counterfactual for any arbitrary game state in a sufficiently complex game. In contrast, our counterfactual state explanation generates the game frame on the fly, and even though it is not perfectly faithful to the game, it has sufficient fidelity to give meaningful insight to participants.

The inconsistency in quality of the NNCEs likely contributed to the confusion of participants. In the retrospective review from participants who were given the Nearest Neighbor counterfactual explanations, they frequently self-reported to being confused by the highlights or the counterfactual itself. For instance, when asked if the highlights helped make their decision:

\begin{displayquote}\textbf{Participant 17 in Nearest Neighbor condition: }
``Sometimes, but it got confusing sometimes as I couldn't tell what highlight belonged to what image, so I couldn't get the AI's thoughts from it''
\end{displayquote}

Similarly, when asked if the counterfactual state image helped make their decision:
\begin{displayquote}\textbf{Participant 26 in Nearest Neighbor condition: }
``No, I was not sure how it related to the decision to move or shoot.''
\end{displayquote}


The NNCEs seemed to be useful to only a small group of participants, as $17\%$ of the participants were able to accurately diagnose the specific flaw in the flawed agent, up from none ($\alpha=0.05$, p-value = 0.014, Pearson's Chi Square test). This percentage is slightly higher than 12.5\%, which is the probability of correctly guessing the flawed agent and correctly guessing the correct flaw purely by chance. 

%% file: discussion.tex


In summary, participants in our first study found our counterfactual state explanations to generate game frames that were close in terms of fidelity though not perfectly so. In our second user study, these counterfactual state explanations were of sufficient fidelity that $90\%$ of our participants could use them to identify which agent was flawed. $60\%$ of the participants were able to use our counterfactual state explanations to perform the more challenging task of diagnosing the specific flaw. During this study, participants specifically mentioned the highlights and summary chart as being particularly useful in their decision-making, thereby suggesting that these visual elements greatly enhance counterfactual explanations.

Our counterfactual state explanations were also much more effective in our user study than the nearest neighbor baseline. Participants using the counterfactual state explanations were much more successful at identifying the flawed agent as well as the specific flaw than participants using the NNCEs. In addition, even though the NNCEs were 100\% faithful to the game, they were not always close to the query state. Participants found that our counterfactual state explanations, which produced images that were "close" to the original query state, were more insightful despite not having 100\% fidelity. Our study also indicated that ``No explanation is better than a bad explanation'' as participants using the NNCEs were often confused and the number of participants correctly identifying the flawed agent actually decreased after seeing the NNCE. 

There are a few issues with our approach that remain an open area of investigation. First, our deep generative approach adds some artefacts when creating counterfactual states, which impacts the faithfulness of our explanation. Empirically, we found most artefacts were minor, such as blurry images, and did not seem to be a major roadblock for our participants. One of the more noticeable artefacts is how small objects, such as the shot in space invaders, occasionally disappear. While this issue is somewhat alleviated with max loss clipping, small objects are difficult to preserve in the counterfactual generation processes. These small objects, however, could be important for other domains (e.g. Pong). It is likely that some of these artefacts could be fixed by training longer, with more data, and with better architectures. 
This problem also raises an open question in representation learning about preserving small, but important, objects in images. 

A second issue is how to select query states from a replay such that the counterfactual states, and actions, provide the most insight to a human. Our criterion was based on heuristics and a deeper investigation is needed as other criteria could be used, such as those used by other methods for selecting key moments \citep{Amir18,Sequeira20}. Moreover, we chose the counterfactuals to present to the participants using a heuristic rather than allowing the participants to interactively explore the space of counterfactuals. We made this choice because many of the counterfactual actions result in no change to the image and users need more guidance as to which counterfactual actions and states are useful. We recognize that this choice directly affects the diversity among the counterfactuals that the participants see, and may hinder in the building of a sufficiently well-rounded mental model \citep{DiverseCounterfactuals2019}. 

Another area for future work is in choosing the way a counterfactual state is generated. Our counterfactual state generation was based on finding a state $\bm{s'}$ that was minimally changed (in the latent space) from the query state $\bm{s}$ that would result in a different action $a'$ from $a$. The reason for the minimal change was to identify the necessary aspects of a state that would produce the action $a'$,  without distracting the user with other irrelevant elements in the image. This minimal change criterion is similar to approaches used by other recent methods for generating counterfactuals, such as the minimal-edit approach for replacing regions in an image \citep{goyal_countervis_2019} and the search for smallest deletion / supporting regions \citep{chang2019explaining}. However, we could use other criteria besides minimal change to define the space of modifications to the query state. For instance, we could permit changes that lead to specific properties on future time steps or allow the user to help define the space of changes allowed. 

Finally, we recognize that our findings are specific to the visual input environment of Atari, and the success of a generative deep learning method for producing counterfactuals in other visual environments is an open question. In particular, the fidelity of the counterfactual states depends on the amount of training data available and the ability of the deep neural network to capture salient aspects of the images from that domain. 
While the primary application of our work is for Atari-like domains, more sophisticated auto-encoding training methods have been shown to produce high quality images in more visually rich environments \citep{nie2020semi}.
Therefore, we believe that there are some findings in our study that could be more broadly applicable. The general framework of our explanation, namely presenting original--highlights--counterfactual could be effective in many domains. Moreover, our results also indicate that perfect fidelity may not be necessary. Counterfactual images with sufficient fidelity could give enough insight in other domains and even the highlights by themselves might be sufficiently insightful for other visual environments.






%% file: conclusion.tex
We introduced a deep generative model to produce counterfactual state explanations as a way to provide insight into a deep RL agent's decision making. The counterfactual states showed what minimal changes needed to occur to a state to produce a different action by the trained RL agent. Results from our first user study showed that these counterfactual state explanations had sufficient fidelity to the actual game. Results from our second user study demonstrated that despite having some artefacts, these counterfactual state explanations were indeed useful for identifying the flawed agent in our study as well as the specific flaw in the agent. In comparison, the nearest neighbor counterfactual explanations confused participants and resulted in fewer participants identifying the correct agent after they were shown the explanation. Furthermore, only a small proportion of participants were able to identify the specific flaw. Our study also demonstrated that the highlights and summary table were important elements to accompany counterfactual explanations. 

Our results suggest that perfect fidelity may not be necessary for counterfactual state explanations to give non-machine learning experts sufficient understanding of an agent's decision making in order to use this knowledge for a downstream task.  While our study focused on Atari agents, we believe this approach is promising and could apply more broadly to domains beyond Atari with more complex visual input though more investigation is needed. Moreover, using counterfactual state explanations in conjunction with other established and complementary explanation techniques could form a formidable toolset to help non-experts understand decisions made by deep RL agents.





%% file: appendix.tex
\section{Tuning the $\lambda$ Parameter}

\begin{figure*}[htb]
    \centering
    \includegraphics[width=.19\linewidth]{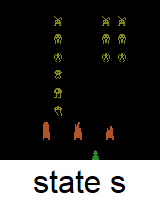}
    \includegraphics[width=.19\linewidth]{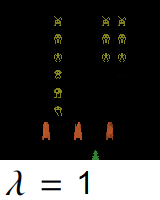}
    \includegraphics[width=.19\linewidth]{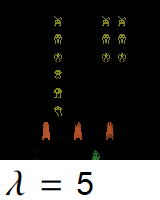}
    \includegraphics[width=.19\linewidth]{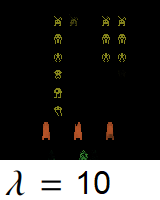}
    \includegraphics[width=.19\linewidth]{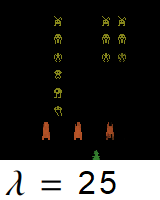}
    \includegraphics[width=.19\linewidth]{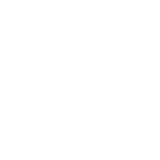}
    \includegraphics[width=.19\linewidth]{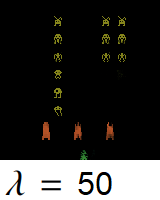}
    \includegraphics[width=.19\linewidth]{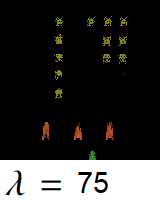}
    \includegraphics[width=.19\linewidth]{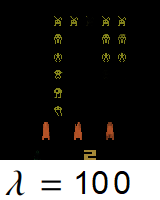}
    \includegraphics[width=.19\linewidth]{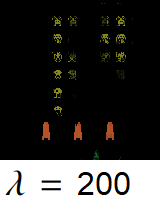}
    \caption{ An example of different models trained with varying $\lambda$ parameters for the normally trained agent. The original state with action $a =$ \textit{MoveLeftAndShoot} for which to generate a counterfactual is shown in the top left, with the rest of the images being counterfactual states where the agent would take counterfactual action $a' =$ \textit{Fire}.}
    \label{fig:lambda_tuning}
\end{figure*}

\label{lambda_tuning}


Figure \ref{fig:lambda_tuning} shows the effects of varying the $\lambda$ parameter. As $\lambda$ increases, so does the amount of change in the counterfactual state, with low $\lambda$ values causing nearly imperceptible changes and high $\lambda$ values producing distorted, low quality states.
From our first user-study, we found that non-experts were able to clearly identify poor fidelity images, caused by $\lambda$ parameters that were too high. Given a set of images produced by different $\lambda$ values, we feel that finding a "sweet spot" between too high and too low should be manageable for a non-expert viewer as there is a fairly wide range of $\lambda$ values that produce reasonably high quality counterfactuals. Automating the process of selecting a $\lambda$ for high fidelity counterfactual production is beyond the scope of this work, but is an area of future interest.

\section{Ablation Experiments for the Counterfactual State Neural Network Architecture}

\begin{table}[h]
\centering
\begin{tabular}{|l|l|l|l|l|l|l|l|l|l|l|}
\hline
Ablation experiment & 1  & 2  & 3  & 4  & 5  & 6  & 7  & 8  & 9  & 10 \\ \hline
$E(s)$                &    &    &    &    &    & \checkmark & \checkmark & \checkmark & \checkmark & \checkmark \\ \hline
$z$                   &    & \checkmark & \checkmark &    &    &    & \checkmark & \checkmark &    &    \\ \hline
$z_w$                &    &    &    & \checkmark & \checkmark &    &    &    & \checkmark & \checkmark \\ \hline
 $\bpi(A(\bs))$           & \checkmark &    & \checkmark &    & \checkmark & \checkmark &    & \checkmark &    & \checkmark \\ \hline
\end{tabular}
\caption{An overview of what elements are given as input to the generator for each ablation experiment.}
 \label{table:ablations}
\end{table}

\label{ablation_descriptions}

\begin{figure*}[htb]
    \centering
    \includegraphics[width=.19\linewidth]{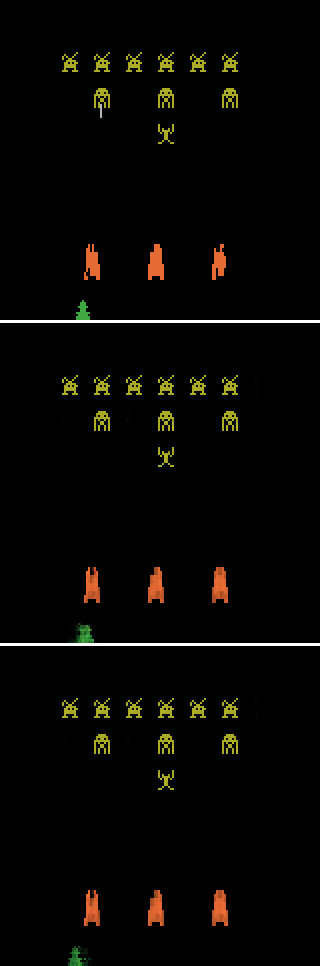} 
    \includegraphics[width=.19\linewidth]{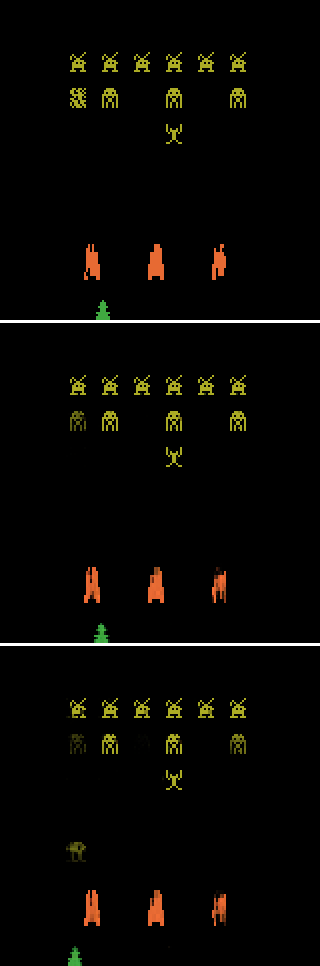}
    \includegraphics[width=.19\linewidth]{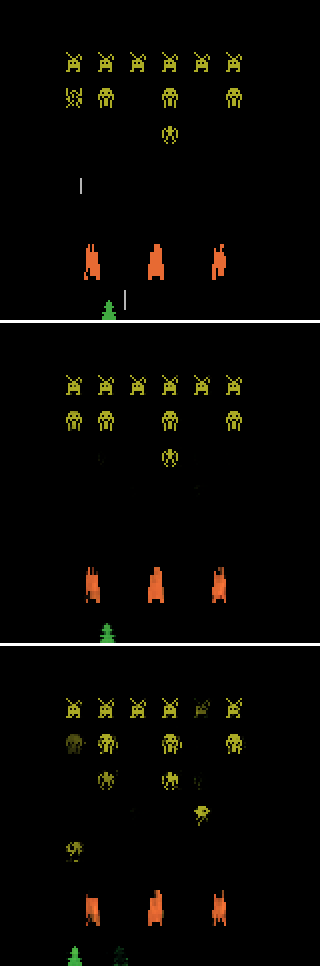}
    \includegraphics[width=.19\linewidth]{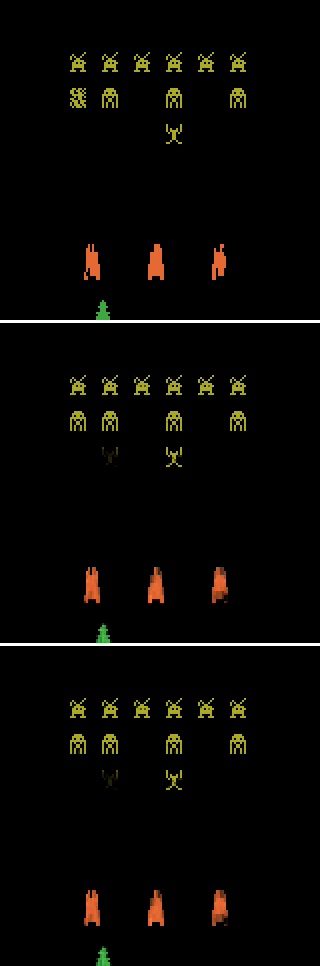}
    \includegraphics[width=.19\linewidth]{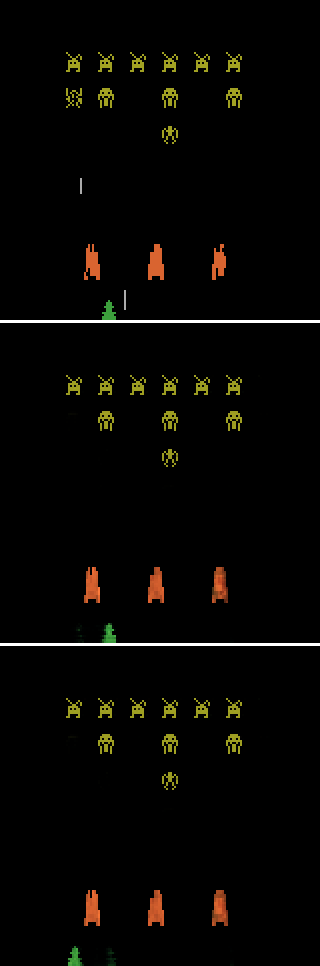}
    \caption{From left to right: ablations 6 - 10. In each column, the top image is an original state where $a =$ \textit{MoveRight}, the middle image is an auto-encoded reconstruction, and the bottom is a counterfactual state where $a'=$ \textit{MoveRightAndFire}.}
    \label{fig:ablations2}
\end{figure*}

In this section, we describe many different ablation experiments using the neural network architecture described in Section \ref{sec:method_arch} for our counterfactual state explanations. These ablations illustrate how each component in our architecture is necessary to achieve high quality counterfactual images. A generator is always used with MSE reconstruction loss, but what we pass into the generator changes for each ablation experiment. We provide an overview of the different ablations in Table \ref{table:ablations} and Figures \ref{fig:ablations1} and \ref{fig:ablations2} contain images which are representative of the issues for each ablation experiment. Next, we discuss each ablation experiment in detail:

\begin{figure*}[h]
    \centering
    \includegraphics[width=.19\linewidth]{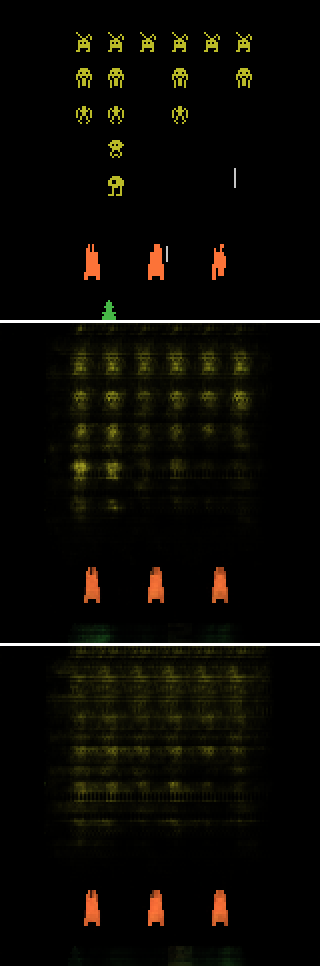}
    \includegraphics[width=.19\linewidth]{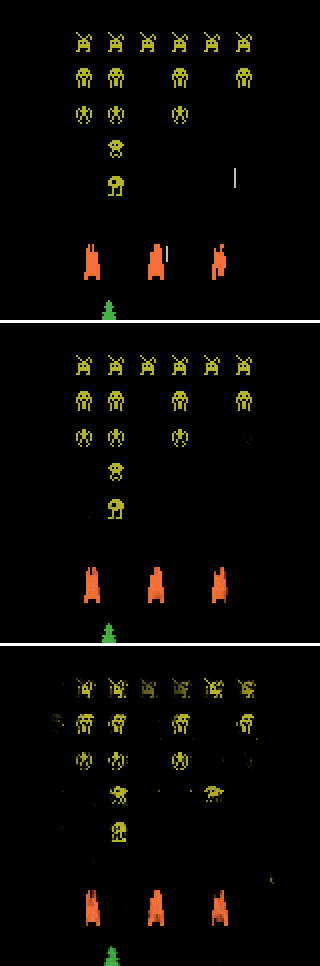}
    \includegraphics[width=.19\linewidth]{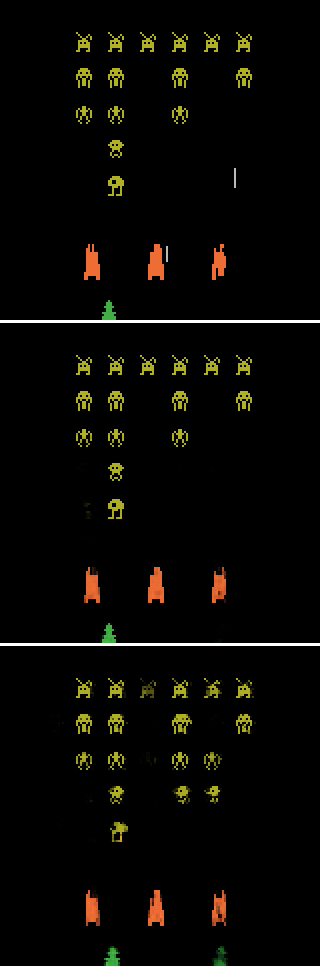}
    \includegraphics[width=.19\linewidth]{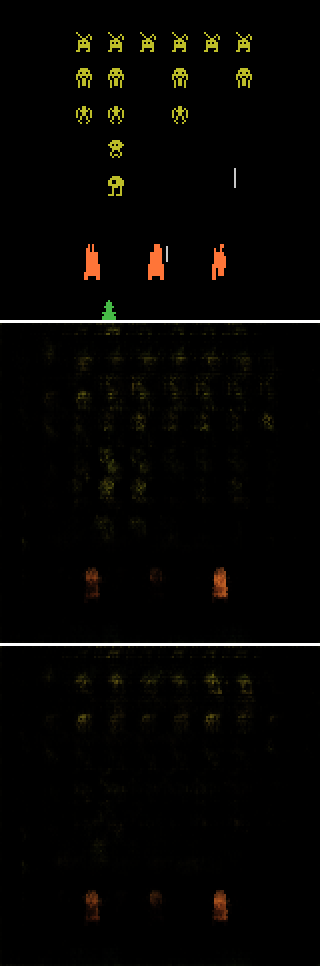}
    \includegraphics[width=.19\linewidth]{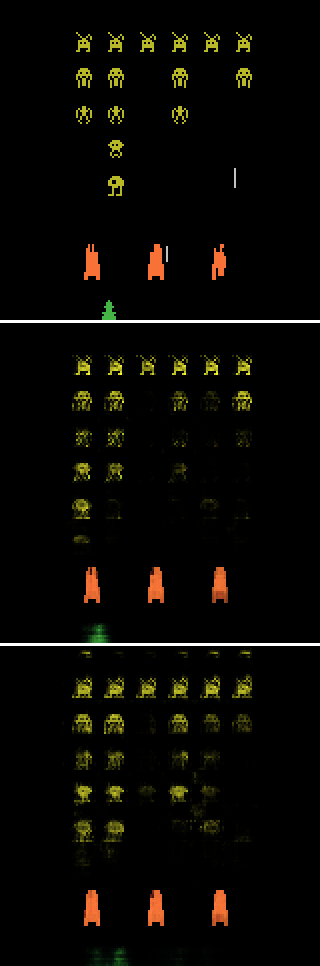}
    \caption{From left to right: ablations 1 - 5.  In every column and each row, the top image is the original state where $a =$ \textit{MoveRight}, the center image is the auto-encoded reconstruction of the original state, the bottom image is a counterfactual state where $a'=$ \textit{MoveRightAndFire}.}
    \label{fig:ablations1}    
\end{figure*}
\begin{enumerate}
    \item  We investigate the effect of only using the agent's policy to generate reconstructed states and hand-modifying it to create counterfactual states: removing all parts from our model except the agent and the generator, passing solely the $\bpiz$, where $\bz = A(\bs)$, into the generator. This gives us reconstructed states in the form of $G(\bpiz)$. We modify the policy vector $\bpiz$ by selecting a counterfactual action $a'$, setting $\bpi(\bz,a') = \bpiza * 1.01$, and normalizing the probabilities back to 1. This hand modification is clearly not representative of the agent. As shown in figure \ref{fig:ablations1}, the reconstructed and counterfactual states are extremely low quality. 
    \item We investigate the effect of only using the agent's learned representation to generate both reconstructed states and counterfactual states. We removed all parts from our model except the agent and the generator, passing $\bm{z}$ into the generator. This gives us reconstructed states in the form of $G(\bz)$ and counterfactual states by modifying $\bz$ with gradient descent as described in Section \ref{sec:generating_counterfactuals} to get a $\bm{z}^*$. As shown in figure \ref{fig:ablations1}, the counterfactual states are quite unrealistic, but surprisingly the reconstructed states are accurate. 
    \item We removed all parts from our model except the agent and the generator, this time passing both $\bz$ and $\bpiz$ into the generator. This gives us reconstructed states in the form of $G(\bz,\bpiz)$ and counterfactual states by modifying $\bz$ with gradient descent as described in Section \ref{sec:generating_counterfactuals} to get a $\bm{z}^*$. As shown in figure \ref{fig:ablations1}, the counterfactual states are quite unrealistic, but surprisingly the reconstructed states are accurate.
    \item We investigate using only the Wasserstein auto-encoder. Here we pass only $\bzw$ into the generator, where $\bzw$ is the latent representation of the state in Wasserstein space $\bzw = E_w(A(\bs))$. This gives us reconstructed states in the form of $G(\bzw)$ and counterfactual states by modifying $\bzw$ with gradient descent as described in Section \ref{sec:generating_counterfactuals} to get a $\bm{z_w}^*$. As shown in figure \ref{fig:ablations1}, both the reconstructed and counterfactual states are quite unrealistic.
    \item We removed all parts from our model except the agent, the Wasserstein auto-encoder, and the generator. Here we pass both $\bzw$ and $\bpiz$ into the generator. This gives us reconstructed states in the form of $G(\bz,\bpiz$) and counterfactual states by modifying $\bzw$ with gradient descent as described in Section \ref{sec:generating_counterfactuals} to get a $\bm{z_w}^*$. As shown in Figure \ref{fig:ablations2}, both the reconstructed and counterfactual states improve relative to the previous ablation, but are still quite unrealistic.
    \item
    
    Here we investigate the effect of keeping the encoder and discriminator, but hand-modify the policy input to the generator instead of using the Wasserstein auto-encoder or gradient descent. The input to the generator is equivalent to our work described in section 3. We hand-modify the policy vector $\bpiz$, by selecting a counterfactual action $a'$, setting $\bpi(\bz,a') = \bpiza * 1.01$, and normalizing the probabilities back to 1. These hand modification may, or may not, be representative of what the agent does. As shown in figure \ref{fig:ablations1}, the states have the same generated quality as our method and the counterfactual state has a small, but meaningful change.  
    
    \item
    
    This ablation is similar to the previous ablation, but instead of passing the policy vector $\bpiz$ to the generator, we input the agent's latent space $\bz$.
    As with previous ablations, we generate counterfactual states by modifying $\bz$ with gradient descent as described in Section \ref{sec:generating_counterfactuals} to get a $\bm{z}^*$. As shown in figure \ref{fig:ablations2}, the states have a decent quality, but the counterfactual states have relatively large changes and a couple of artifacts.  
    \item 
    
    Similar to the previous ablation, but instead of passing just $\bz$, we pass in both the policy vector $\bpiz$ and $\bz$ to the generator.
    As with previous ablations, we generate counterfactual states by modifying $\bz$ with gradient descent as described in Section \ref{sec:generating_counterfactuals} to get a $\bm{z}^*$. As shown in figure \ref{fig:ablations2}, the states are better quality than just passing in $\bm{z}^*$, but the counterfactual are lower quality than our method.  
    \item 
    
    We add back in the Wasserstein auto-encoder to the previous ablation. Instead of passing in the agent's latent space $\bz$ to the generator, we pass in the Wasserstein representation $\bzw = E_w(A(\bs))$.
    As described in \ref{sec:generating_counterfactuals}, we generate counterfactual states by modifying $\bzw$ to get a $\bm{z_w}^*$. As shown in figure \ref{fig:ablations2}, the states are high quality, but the counterfactual states typically have no changes.  
    \item This experiment is an ablation in the sense that we remove the disconnection between the generation and $\bzw$. In other words, we take our original method and add $\bzw$ as input to the generator. When counterfactual states are generated, $\bm{z_w}^*$ is passed into the generator along with $E(\bs)$ and $\bm{\pi}(\bm{z_w}^*)$. As shown in figure \ref{fig:ablations2}, the states are high quality and the counterfactual states are interesting. We were not able to find a difference in quality for generated states between this ablation and our method. Since this ablation is more complex, and requires more parameters, we decided not to use it for our purposes.
\end{enumerate}

\section{Details for User Study 2}
In this section, we provide further details on the second user study. Specifically, we include the tutorial script and the images used in the second user study.

\subsection{User Study Tutorial Script}
\begin{figure}[t]
    \centering

    \subfloat{
    \includegraphics[width=.25\linewidth]{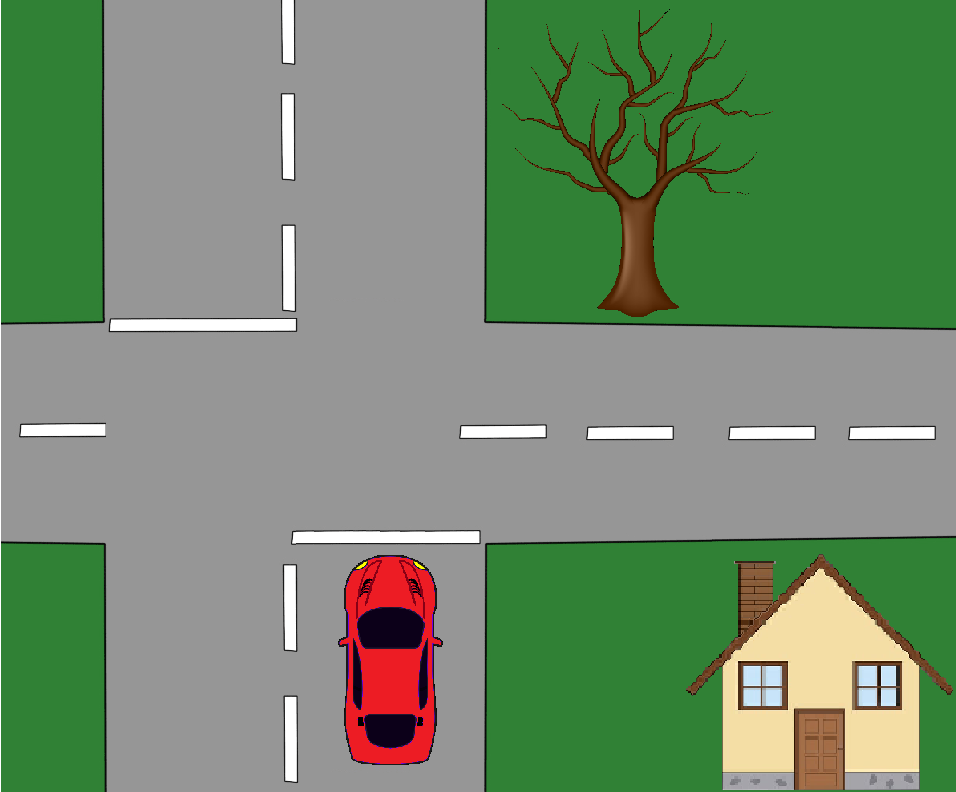}
    \includegraphics[width=.25\linewidth]{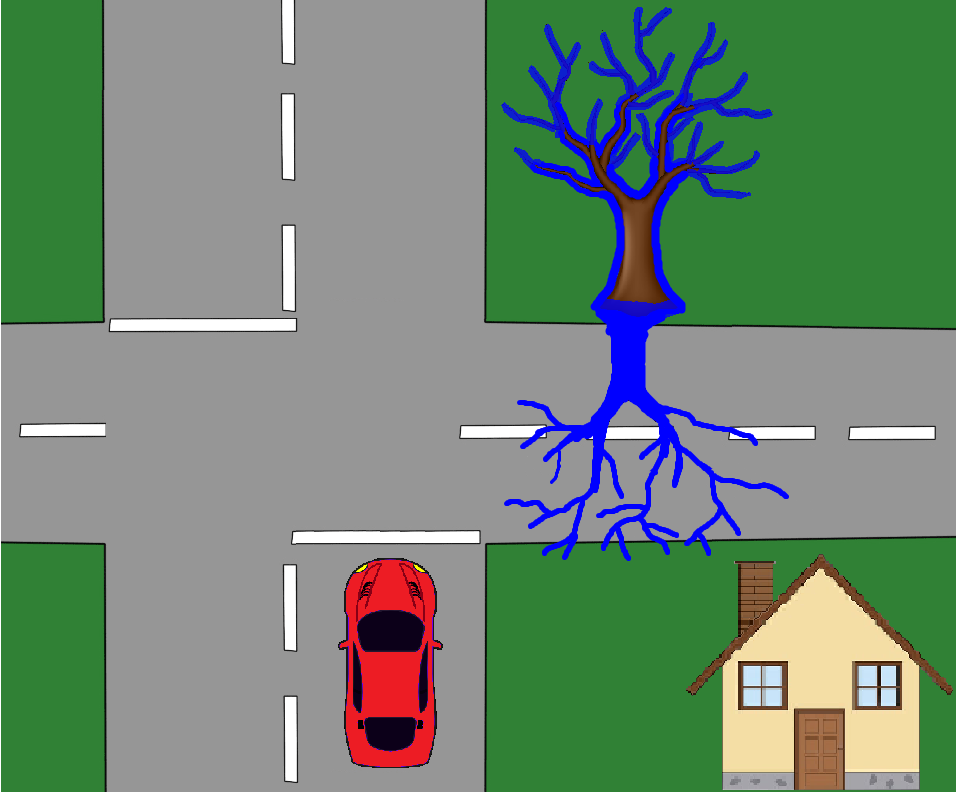}
    \includegraphics[width=.25\linewidth]{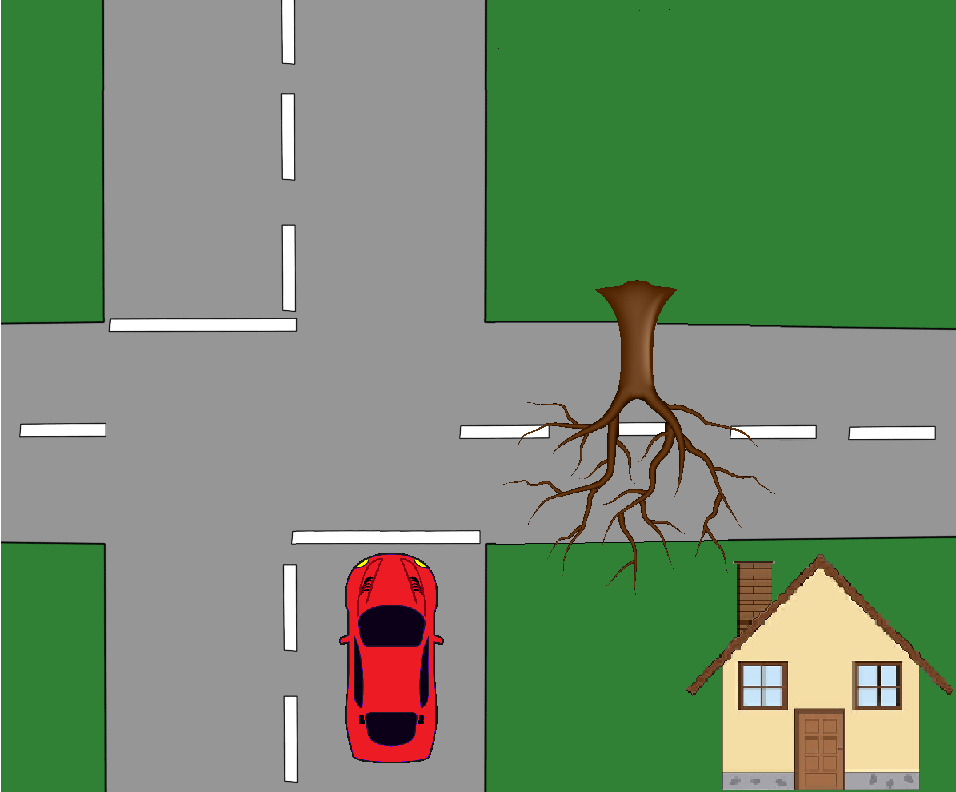}
    }
    
     \subfloat{
    \includegraphics[width=.25\linewidth]{images/tutorial_script/tutorial_original.png}
    \includegraphics[width=.25\linewidth]{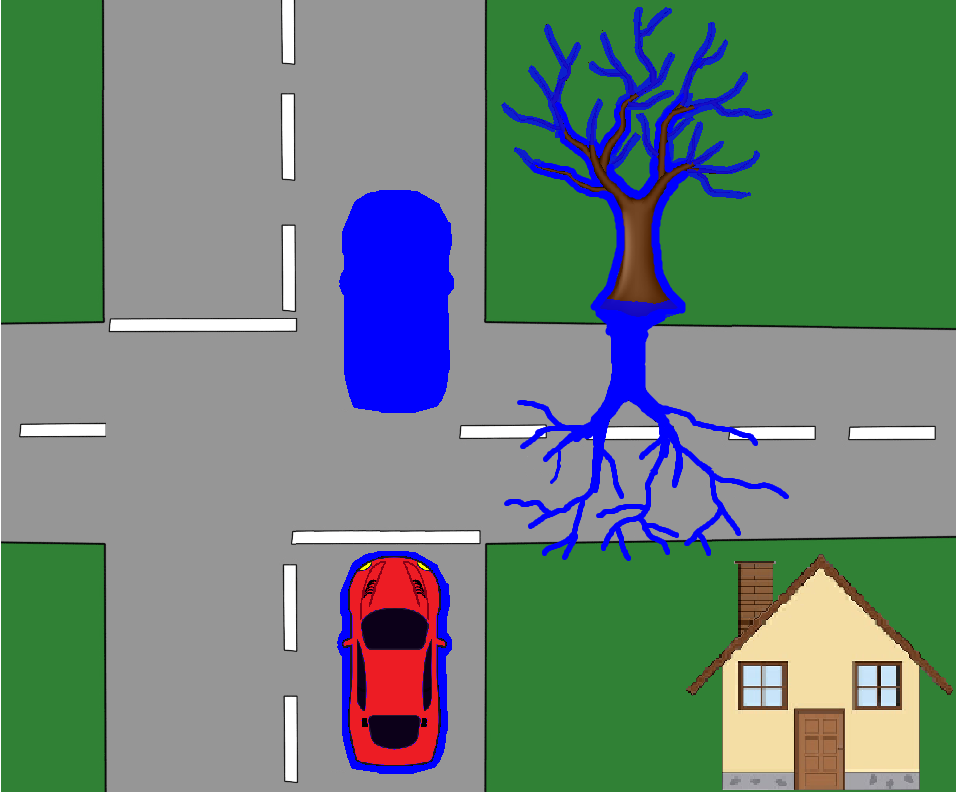}
    \includegraphics[width=.25\linewidth]{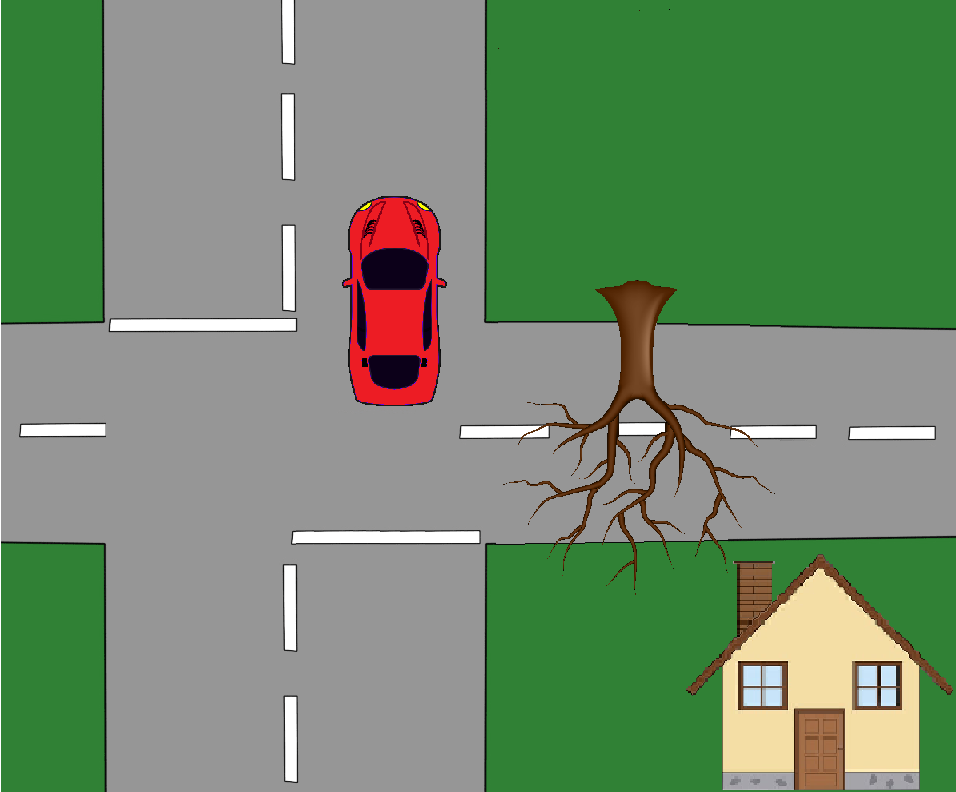}
    }
    
    \caption{
    The user study tutorial examples used to describe counterfactual states, where the top row of images is one potential counterfactual explanation and the bottom is another.  Query state with action $a=\text{TurnRight}$ where a self-driving car is taking you home (\textbf{left}), counterfactual state where action $a'=\text{GoStraight}$ (\textbf{right}), and the highlighted difference (\textbf{center}).
         }

    \label{fig:tutorial_car_counterfactual}%
    
\end{figure}

\label{sec:script}
In this tutorial we will introduce you to the tool for finding the malfunctioning AI.  This tool shows the AIs response to specific “What if” questions. Both the functioning and malfunctioning AI provide answers to the “What if” questions.

For this study, we have selected 20 different screen shots from the videos. After learning how to use the tool, you will examine the selected screen shots to collect data on the two AIs. The identity of the AIs will remain anonymous until the final evaluation.
At this time, please click the checkbox, then the continue button.

For each selected screen shot, you will see three images arranged in a table. We will now go over how the table is arranged.
Please click Next.

The first image is a screen shot from the original videos. 
Please click Next.

In this column, you will also see context for the original screen shot with a short gif. 
Please click Next.

Click on the image to change it into a gif. The gif shows the three previous game states.
Then click again to return image.
In the column, you will also see the original action the AI decided it will take at that moment in the video. 
Please click Next.

In this example, the AI originally decided it would take the "shoot" action. We then asked the AI, "What would the current screen need to look like for you to perform the "move right" action?”
To answer this question, the AI will only evaluate the current moment in the game, not the past or the future.
Please click Next.

For a more concrete example, consider the following. Imagine there is a red self-driving car that is taking you home. It approaches an intersection and it wants to turn right to take you to your destination.
(Reveal Figure \ref{fig:tutorial_car_counterfactual} top-left)

Now imagine a situation where the red car would choose to go straight instead of turning right.
There are various reasons why this could happen. One example is if the brown tree fell over and blocked the road.
(Reveal Figure \ref{fig:tutorial_car_counterfactual} top-right)

In this example, an answer to a question of “what would need to change” right now for the car to choose go straight at this intersection (point to left image),
would be “the brown tree fell over which blocks the right turn”
(point to right image),
Is that clear? 
		
Excellent. Now in the examples you will look at, the AI will answer the question of “what needs to change” by responding with 2 images. 
Please click Next.

The first response is the changed state. This response shows the smallest amount of change in the game to take the different action of “move right”. Back to the car example, if the original image was the intersection (point to left image),
the following response image would be the intersection with the fallen brown tree
(point to right image),
Please click Next.

In the third column, note how the game has subtly changed in two ways: the ship is under the barrier and the barrier is fully armored.
Please click Next.

The second AI response is image highlights, which takes the original screenshot and adds blue highlights to the changes. This response shows where the AI is looking for change to occur. Using the car example, this response would look like the original intersection with a blue highlight where the brown tree has moved.
(Reveal Figure \ref{fig:tutorial_car_counterfactual} top-center)
Is that clear?

Excellent. It is also possible for multiple objects to influence an AIs decision. 
(Reveal Figure \ref{fig:tutorial_car_counterfactual} bottom-left).
In this example, two things influence the red self-driving cars decision to take the move straight action. The first is: if brown tree has fallen over, but also if the red car’s position changed such that it is passed the intersection.
(Reveal entirety of Figure \ref{fig:tutorial_car_counterfactual}).
The highlights for this example show both the red car and the brown tree highlighted in blue.
Is this second example clear?
Excellent. Let us continue with the table and please click Next.

Note how the changed objects are highlighted in blue: the repaired barrier, and the new ship location.
As you are viewing the table for each selected screen shot, you will be asked two questions. The first question is: “what objects in the game do you think the AI pays attention to?”
Please click Next to view this question. You do not need to select an answer for this tutorial.
Do note that you can select more than one checkbox, or no checkboxes at all.

The second question you will be asked is: which AI response or responses did you use for making your decision?
Please click Next. Again, you do not need to answer this for the tutorial.
You will be asked these same questions for every selected screen shot. 

This is the full tool you will be using to analyze each screen shot presented in random order. This section will take about 10 to 15 minutes. For each set of images you will be asked to spend at least 30 seconds. There will be a timer on the screen. 

After you have finished examining the 20 randomized screen shots, you will use the data to complete the second evaluation.  Your results from the tool will be displayed in both a table and a chart. Additionally, we will reveal to you which examples were from AI one and which were from AI two. 

With this information, you will re-answer the question: “which AI is malfunctioning and what objects in the game can it not see?”
And finally, after you have submitted the 2nd evaluation, we ask you to perform a short written reflection.
When you are ready, click “Finish tutorial” to begin viewing the 20 selected screen shots. I will leave the tutorial example on the projector and the car example on the whiteboard. You may begin.

\subsection{Images}
In this section we show a further selection of explanations from our user study. Figures \ref{fig:space_allimgs_normal1} and \ref{fig:space_allimgs_normal2} shows explanations for the normally trained agent for both the counterfactual state explanations and the nearest neighbor counterfactual explanations, sorted by game time step. Figures \ref{fig:space_allimgs_abl1} and \ref{fig:space_allimgs_abl2} similarly shows explanations for the flawed agent. These figures show how the nearest neighbor counterfactual explanations often show the green ship's position changing for the flawed agent, whereas our counterfactual state explanations never change the ship's position.

\newcommand\fw{19}

\begin{figure}[t]
    \centering
    \subfloat{
    \includegraphics[width=.\fw\linewidth]{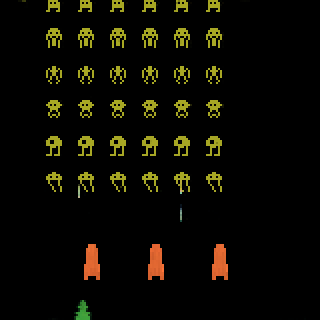}
    \includegraphics[width=.\fw\linewidth]{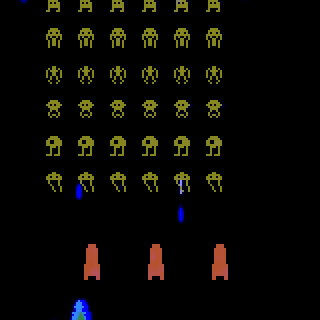}
    \includegraphics[width=.\fw\linewidth]{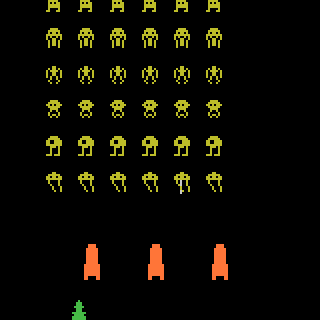} 
    \includegraphics[width=.\fw\linewidth]{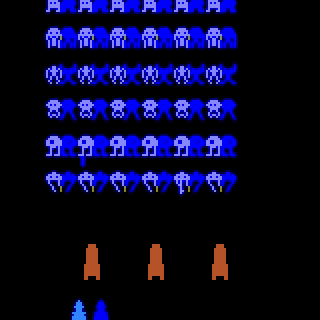} 
    \includegraphics[width=.\fw\linewidth]{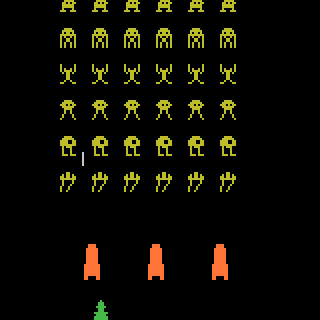} 
    }
    \vspace{-2ex}
    \caption*{$a'=\text{MoveLeft}$  \qquad  $a=\text{MoveRightAndShoot}$,\qquad $a_{nn}'=\text{MoveRight}$}
    \vspace{-2ex}
    \subfloat{
      \includegraphics[width=.\fw\linewidth]{images/user_study/generative_normal_agent/output3_0030_actionRIGHTFIREr5_cf2r2RIGHT.png} 
      \includegraphics[width=.\fw\linewidth]{images/user_study/generative_normal_agent/output2_0030_actionRIGHTFIREr5_cf2r2RIGHT.png} 
      \includegraphics[width=.\fw\linewidth]{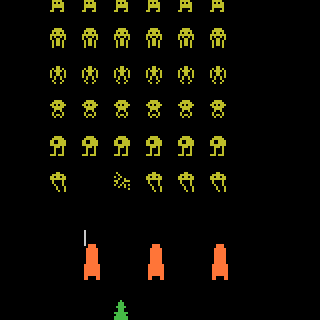} 
      \includegraphics[width=.\fw\linewidth]{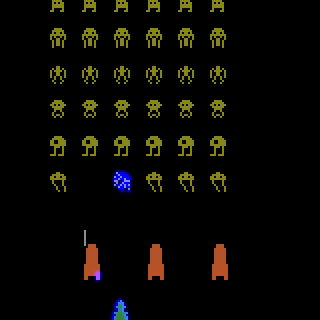} 
      \includegraphics[width=.\fw\linewidth]{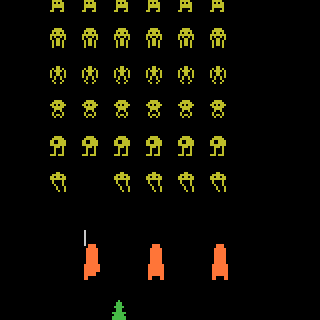} 
    }
    \vspace{-2ex}
    \caption*{$a'=\text{MoveRight}$  \qquad  $a=\text{MoveRightAndShoot}$,\qquad $a_{nn}'=\text{Shoot}$}
    \vspace{-2ex}
     
    \subfloat{
    \includegraphics[width=.\fw\linewidth]{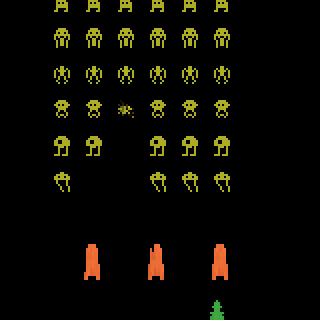} 
    \includegraphics[width=.\fw\linewidth]{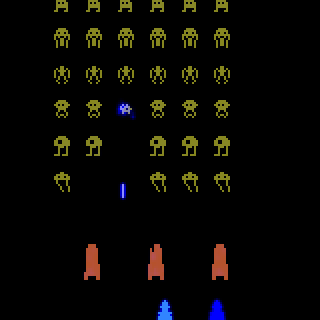}
    \includegraphics[width=.\fw\linewidth]{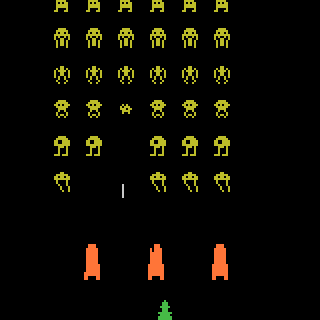}
    \includegraphics[width=.\fw\linewidth]{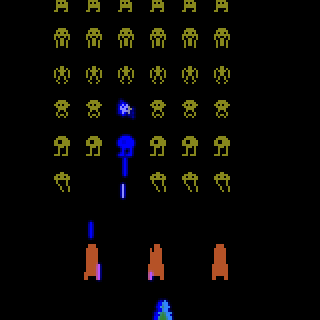}
    \includegraphics[width=.\fw\linewidth]{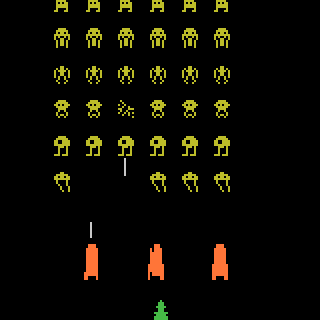}
    }
    \vspace{-2ex}
    \caption*{$a'=\text{MoveRight}$  \qquad  $a=\text{Shoot}$,\qquad $a_{nn}'=\text{MoveRightAndShoot}$}
    \vspace{-2ex}
      \subfloat{
      \includegraphics[width=.\fw\linewidth]{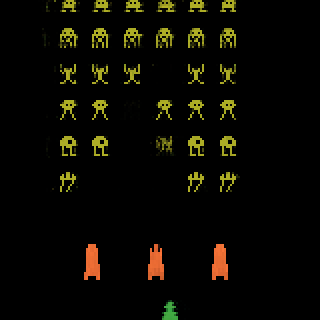}
      \includegraphics[width=.\fw\linewidth]{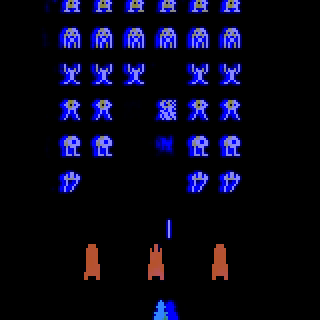}
      \includegraphics[width=.\fw\linewidth]{images/user_study/nn_normal_agent/output1_0065_actionRIGHTFIRE_cf5LEFTFIRE.png} 
      \includegraphics[width=.\fw\linewidth]{images/user_study/nn_normal_agent/output2_0065_actionRIGHTFIRE_cf5LEFTFIRE.png} 
      \includegraphics[width=.\fw\linewidth]{images/user_study/nn_normal_agent/output3_0065_actionRIGHTFIRE_cf5LEFTFIRE.png} 
      }
      \vspace{-2ex}
    \caption*{$a'=\text{MoveLeft}$  \qquad  $a=\text{MoveRightAndShoot}$,\qquad $a_{nn}'=\text{MoveLeftAndShoot}$}
    \vspace{-2ex}
    \subfloat{
    \includegraphics[width=.\fw\linewidth]{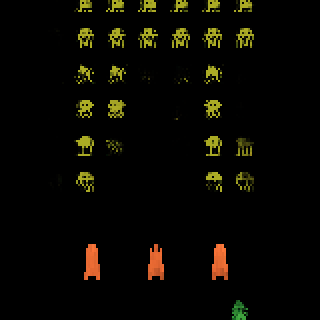} 
    \includegraphics[width=.\fw\linewidth]{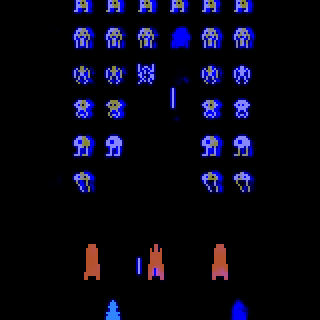}
    \includegraphics[width=.\fw\linewidth]{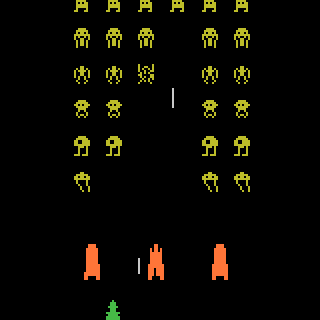}
    \includegraphics[width=.\fw\linewidth]{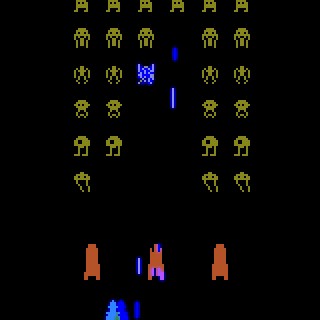}
    \includegraphics[width=.\fw\linewidth]{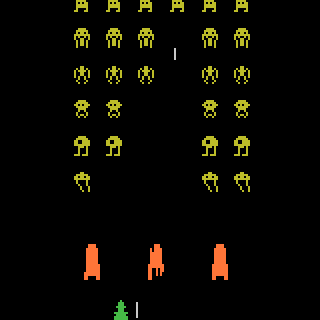}
    }
    \vspace{-2ex}
    \caption*{$a'=\text{MoveLeft}$  \qquad  $a=\text{MoveRightAndShoot}$,\qquad $a_{nn}'=\text{MoveLeftAndShoot}$}
    \vspace{-2ex}

    \caption{
    The first five explanations for the normally trained agent used in the user study.
    (\textbf{Center}) The original state $s$ where the agent took action $a$.
    (\textbf{Left}) The counterfactual state explanations where the agent takes action $a'$.
    (\textbf{Right}) The nearest neighbor counterfactual state where the agent takes action $a_{nn}'$.
    (\textbf{Center Left/Right}) The highlighted difference between the counterfactual state and the original state.
         }

    \label{fig:space_allimgs_normal1}%
\end{figure}

\begin{figure}[t]
    \centering

    \subfloat{
    \includegraphics[width=.\fw\linewidth]{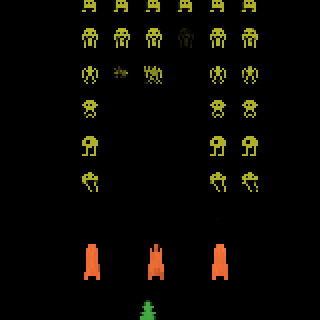} 
    \includegraphics[width=.\fw\linewidth]{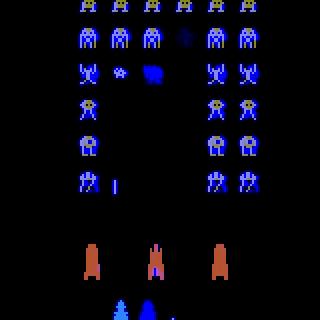} 
    \includegraphics[width=.\fw\linewidth]{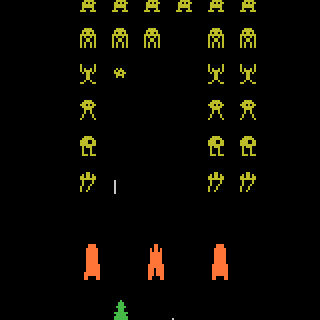} 
    \includegraphics[width=.\fw\linewidth]{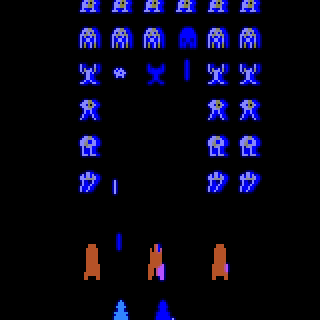} 
    \includegraphics[width=.\fw\linewidth]{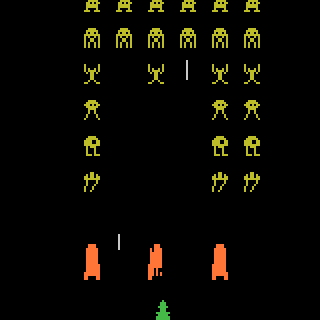} 
    }
    \vspace{-2ex}
    \caption*{$a'=\text{MoveLeft}$  \qquad  $a=\text{MoveRightAndShoot}$,\qquad $a_{nn}'=\text{Shoot}$}
    \vspace{-2ex}
     
      \subfloat{
      \includegraphics[width=.\fw\linewidth]{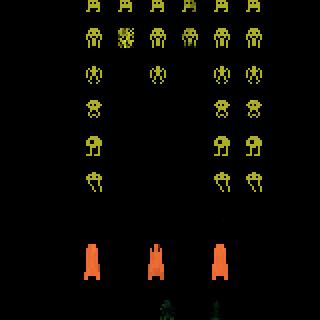} 
      \includegraphics[width=.\fw\linewidth]{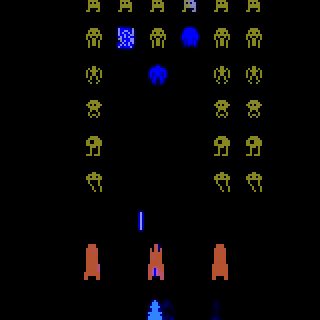} 
      \includegraphics[width=.\fw\linewidth]{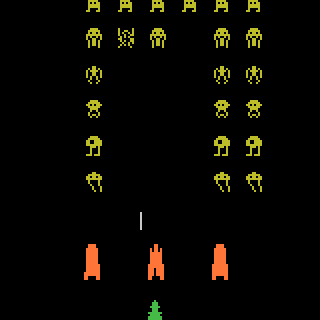} 
      \includegraphics[width=.\fw\linewidth]{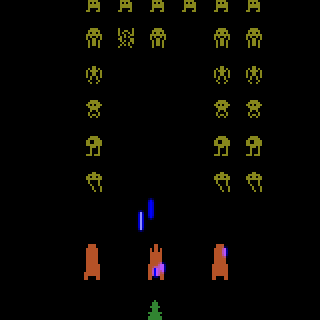} 
      \includegraphics[width=.\fw\linewidth]{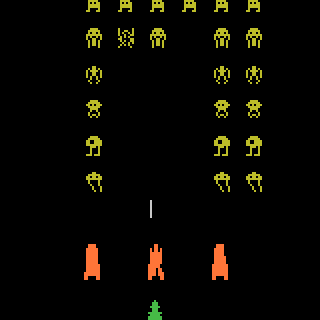} 
      }
        \vspace{-2ex}
    \caption*{$a'=\text{Shoot}$  \qquad  $a=\text{MoveRight}$,\qquad $a_{nn}'=\text{MoveRightAndShoot}$}
    \vspace{-2ex}
      \subfloat{
      \includegraphics[width=.\fw\linewidth]{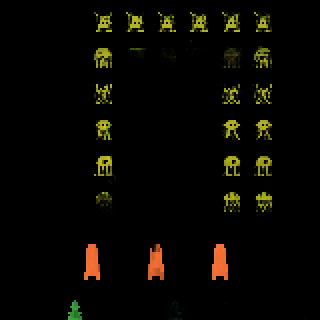} 
      \includegraphics[width=.\fw\linewidth]{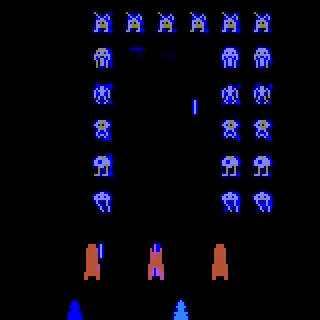} 
      \includegraphics[width=.\fw\linewidth]{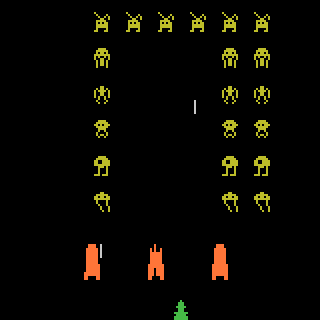} 
      \includegraphics[width=.\fw\linewidth]{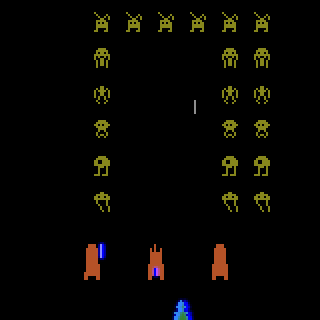}
      \includegraphics[width=.\fw\linewidth]{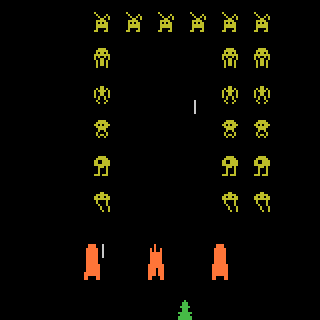}
      }
      \vspace{-2ex}
    \caption*{$a'=\text{Shoot}$  \qquad  $a=\text{MoveRight}$,\qquad $a_{nn}'=\text{MoveRightAndShoot}$}
    \vspace{-2ex}
      \subfloat{
      \includegraphics[width=.\fw\linewidth]{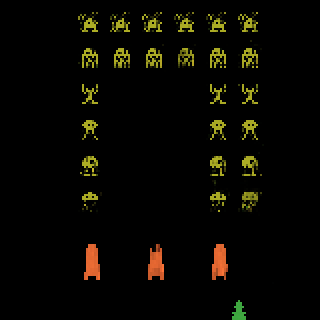} 
      \includegraphics[width=.\fw\linewidth]{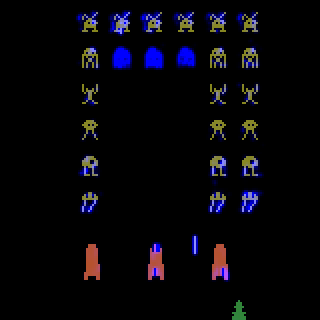} 
      \includegraphics[width=.\fw\linewidth]{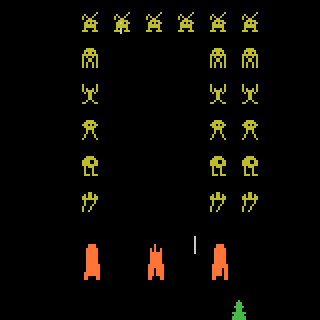} 
      \includegraphics[width=.\fw\linewidth]{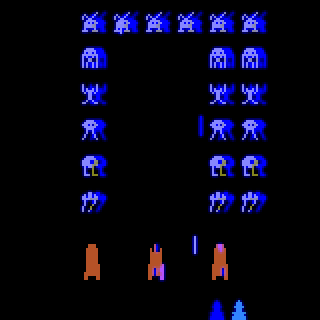} 
      \includegraphics[width=.\fw\linewidth]{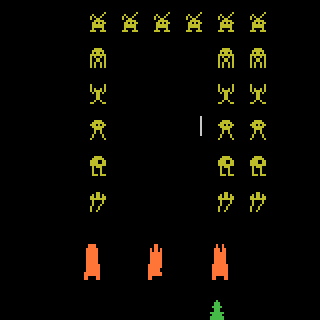} 
      }
      \vspace{-2ex}
    \caption*{$a'=\text{MoveLeft}$  \qquad  $a=\text{MoveRightAndShoot}$,\qquad $a_{nn}'=\text{MoveRight}$}
    \vspace{-2ex}
    \subfloat{
    \includegraphics[width=.\fw\linewidth]{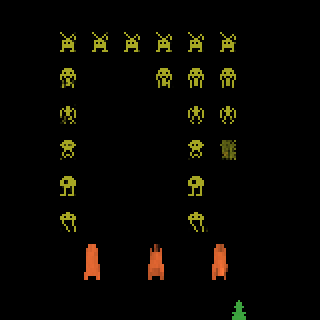} 
    \includegraphics[width=.\fw\linewidth]{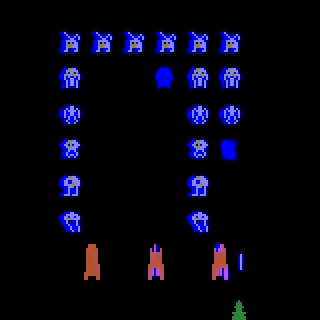} 
    \includegraphics[width=.\fw\linewidth]{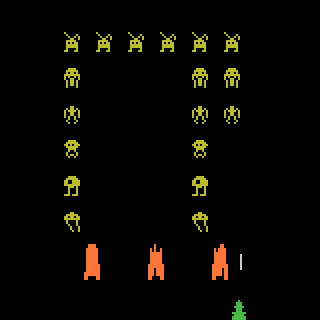} 
    \includegraphics[width=.\fw\linewidth]{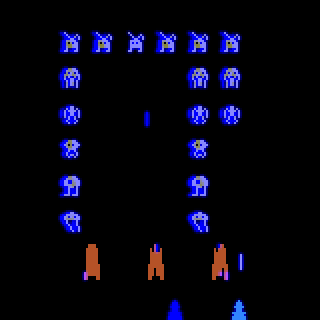} 
    \includegraphics[width=.\fw\linewidth]{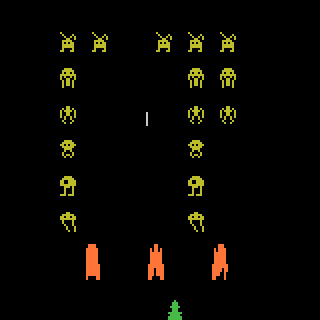} 
    }
    \vspace{-2ex}
    \caption*{$a'=\text{MoveLeft}$  \qquad  $a=\text{MoveRightAndShoot}$,\qquad $a_{nn}'=\text{MoveRight}$}
    \vspace{-2ex}

    \caption{
    Explanations 6 through 10 for the normally trained agent used in the user study.
    (\textbf{Center}) The original state $s$ where the agent took action $a$.
    (\textbf{Left}) The counterfactual state explanations where the agent takes action $a'$.
    (\textbf{Right}) The nearest neighbor counterfactual state where the agent takes action $a_{nn}'$.
    (\textbf{Center Left/Right}) The highlighted difference between the counterfactual state and the original state.
         }

    \label{fig:space_allimgs_normal2}%
\end{figure}

\begin{figure}[t]
    \centering
    \subfloat{
    \includegraphics[width=.\fw\linewidth]{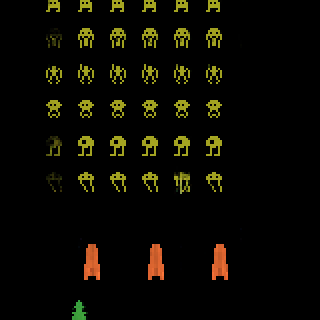}
    \includegraphics[width=.\fw\linewidth]{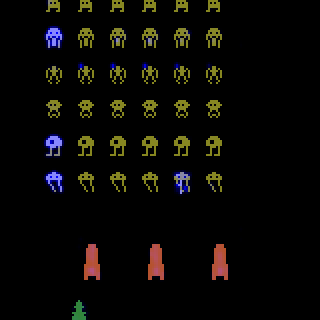}
    \includegraphics[width=.\fw\linewidth]{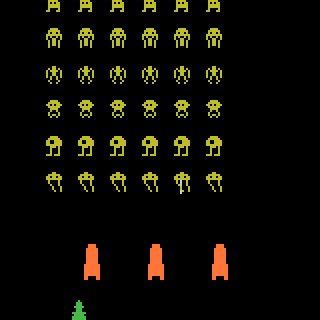}
    \includegraphics[width=.\fw\linewidth]{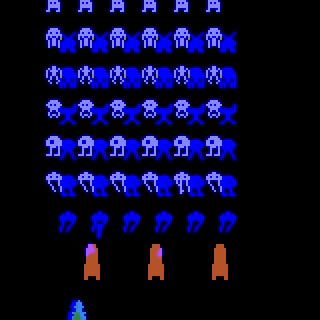}
    \includegraphics[width=.\fw\linewidth]{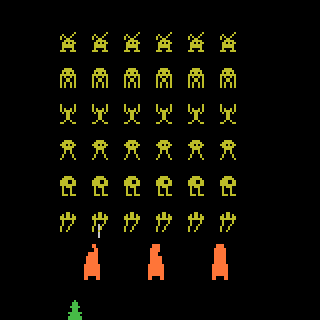}
    }
    \vspace{-2ex}
    \caption*{$a'=\text{MoveRight}$  \qquad  $a=\text{MoveLeft}$,\qquad $a_{nn}'=\text{MoveLeftAndShoot}$}
    \vspace{-2ex}
    \subfloat{
    \includegraphics[width=.\fw\linewidth]{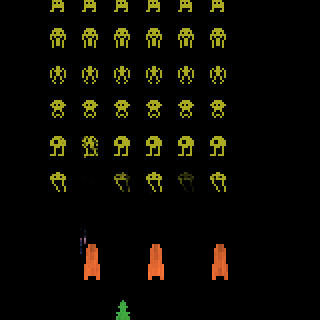}
    \includegraphics[width=.\fw\linewidth]{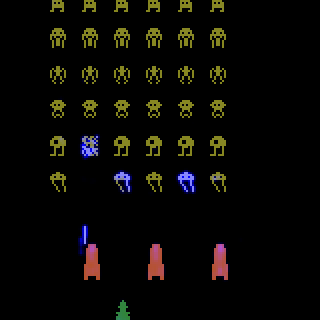}
    \includegraphics[width=.\fw\linewidth]{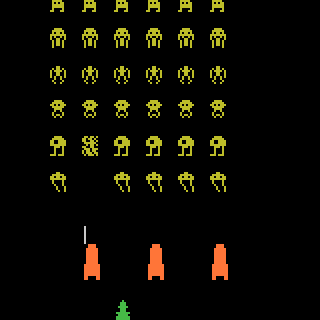}
    \includegraphics[width=.\fw\linewidth]{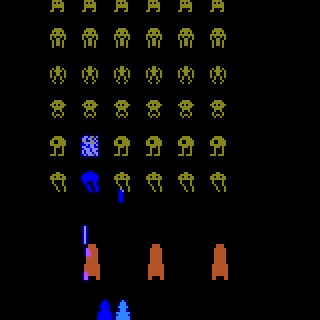}
    \includegraphics[width=.\fw\linewidth]{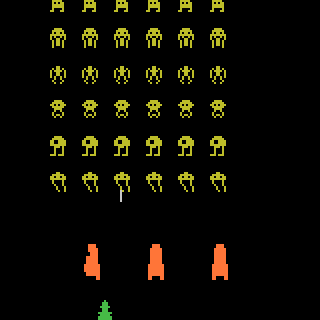}
    }
    \vspace{-2ex}
    \caption*{$a'=\text{MoveRight}$  \qquad  $a=\text{Shoot}$,\qquad $a_{nn}'=\text{StayStill}$}
    \vspace{-2ex}
    \subfloat{
    \includegraphics[width=.\fw\linewidth]{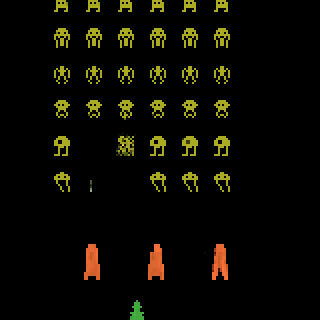}
    \includegraphics[width=.\fw\linewidth]{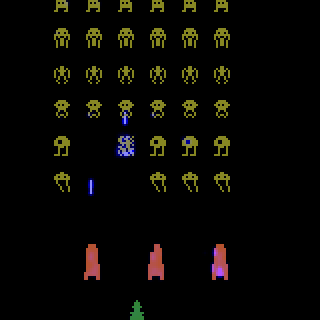}
    \includegraphics[width=.\fw\linewidth]{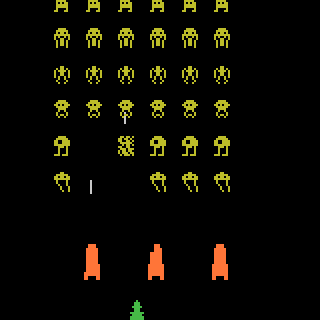}
    \includegraphics[width=.\fw\linewidth]{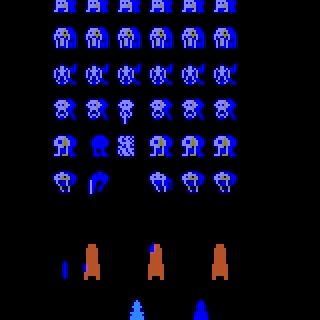}
    \includegraphics[width=.\fw\linewidth]{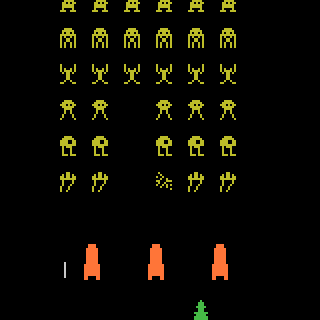}
    }
    \vspace{-2ex}
    \caption*{$a'=\text{MoveLeft}$  \qquad  $a=\text{MoveRightAndShoot}$,\qquad $a_{nn}'=\text{MoveRight}$}
    \vspace{-2ex}
    \vspace{.2ex}
    \subfloat{
    \includegraphics[width=.\fw\linewidth]{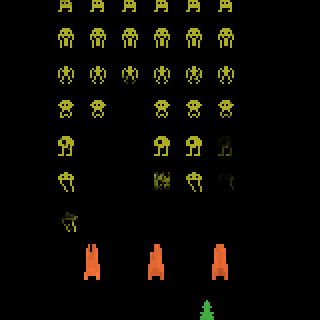}
    \includegraphics[width=.\fw\linewidth]{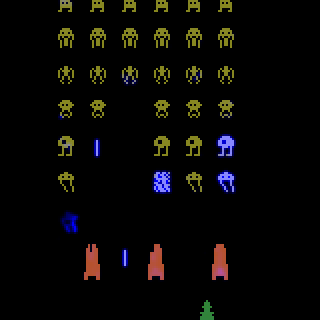}
    \includegraphics[width=.\fw\linewidth]{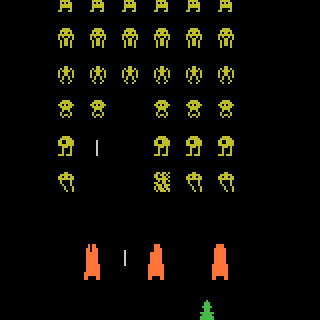}
    \includegraphics[width=.\fw\linewidth]{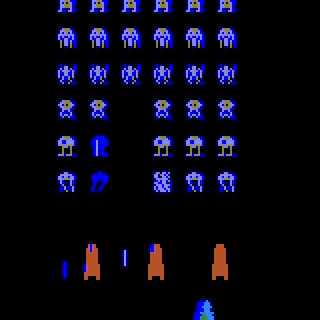}
    \includegraphics[width=.\fw\linewidth]{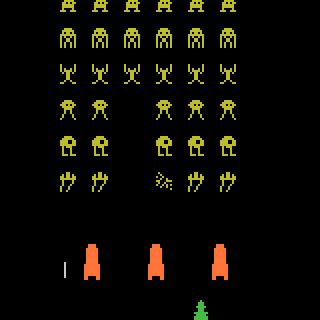}
    }
    \vspace{-2ex}
    \caption*{$a'=\text{MoveLeft}$  \qquad  $a=\text{MoveRightAndShoot}$,\qquad $a_{nn}'=\text{MoveRight}$}
    \vspace{-2ex}
    \subfloat{
    \includegraphics[width=.\fw\linewidth]{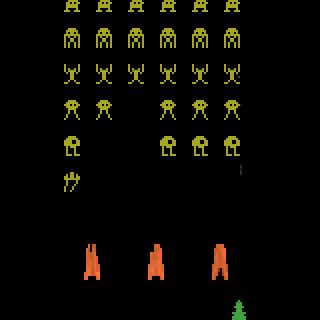}
    \includegraphics[width=.\fw\linewidth]{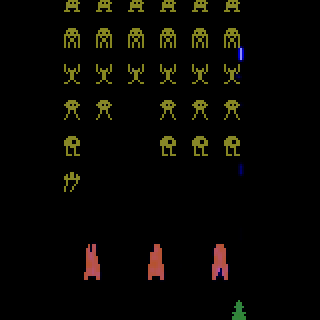}
    \includegraphics[width=.\fw\linewidth]{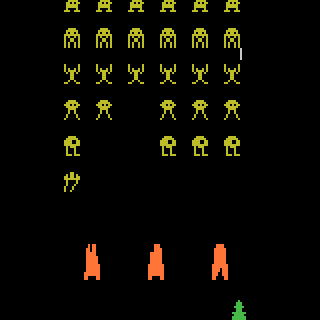}
    \includegraphics[width=.\fw\linewidth]{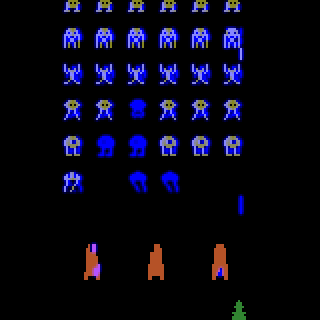}
    \includegraphics[width=.\fw\linewidth]{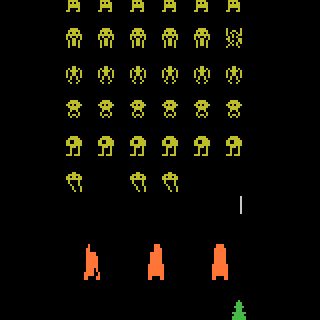}
    }
    \vspace{-2ex}
    \caption*{$a'=\text{MoveRight}$  \qquad  $a=\text{MoveRightAndShoot}$,\qquad $a_{nn}'=\text{Shoot}$}
    \vspace{-2ex}

    \caption{
    The first five explanations for the flawed agent used in the user study.
    (\textbf{Center}) The original state $s$ where the agent took action $a$.
    (\textbf{Left}) The counterfactual state explanations where the agent takes action $a'$.
    (\textbf{Right}) The Nearest Neighbor counterfactual state where the agent takes action $a_{nn}'$.
    (\textbf{Center Left/Right}) The highlighted difference between the counterfactual state and the original state.
         }

    \label{fig:space_allimgs_abl1}%
\end{figure}

\begin{figure}[t]
    \centering

    \subfloat{
    \includegraphics[width=.\fw\linewidth]{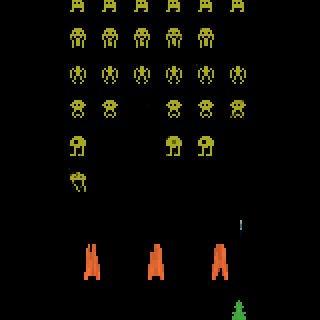}
    \includegraphics[width=.\fw\linewidth]{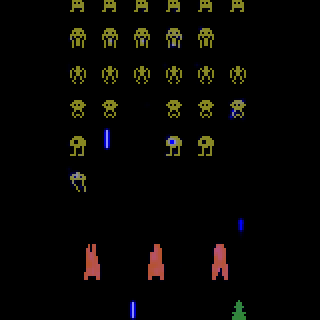}
    \includegraphics[width=.\fw\linewidth]{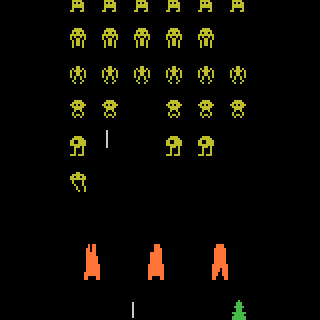}
    \includegraphics[width=.\fw\linewidth]{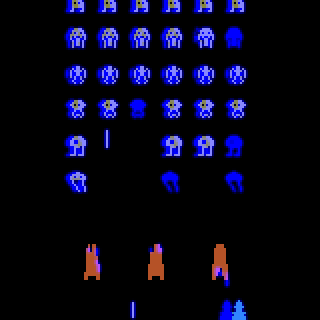}
    \includegraphics[width=.\fw\linewidth]{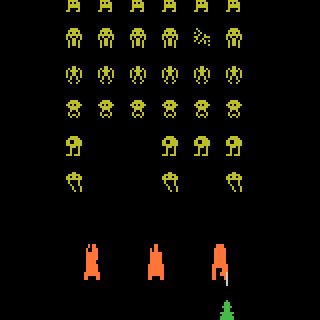}
    }
    \vspace{-2ex}
    \caption*{$a'=\text{MoveRight}$  \qquad  $a=\text{MoveRightAndShoot}$,\qquad $a_{nn}'=\text{Shoot}$}
    \vspace{-2ex}
    
    \subfloat{
    \includegraphics[width=.\fw\linewidth]{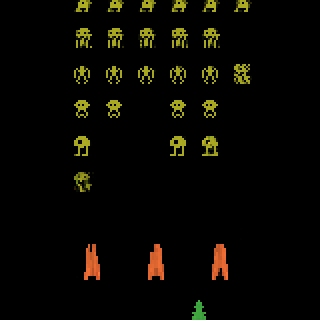}
    \includegraphics[width=.\fw\linewidth]{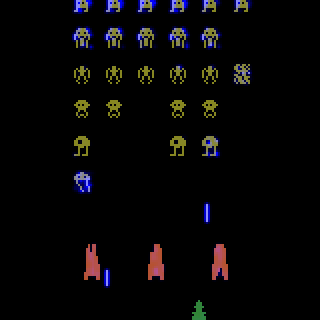}
    \includegraphics[width=.\fw\linewidth]{images/user_study/nn_abl_agent/output1_0086_actionLEFTFIRE_cf3LEFT.png}
    \includegraphics[width=.\fw\linewidth]{images/user_study/nn_abl_agent/output2_0086_actionLEFTFIRE_cf3LEFT.png}
    \includegraphics[width=.\fw\linewidth]{images/user_study/nn_abl_agent/output3_0086_actionLEFTFIRE_cf3LEFT.png}
    }
    \vspace{-2ex}
    \caption*{$a'=\text{MoveRight}$  \qquad  $a=\text{MoveLeftAndShoot}$,\qquad $a_{nn}'=\text{MoveLeft}$}
    \vspace{-2ex}
    \subfloat{
    \includegraphics[width=.\fw\linewidth]{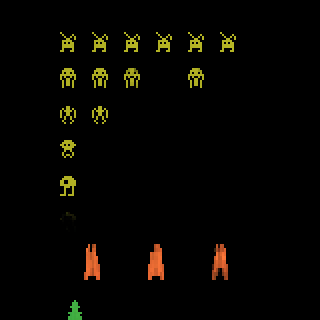}
    \includegraphics[width=.\fw\linewidth]{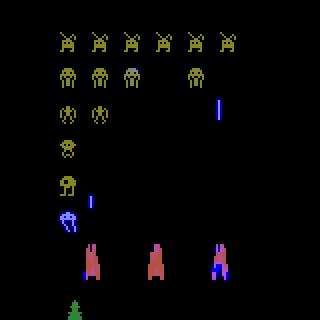}
    \includegraphics[width=.\fw\linewidth]{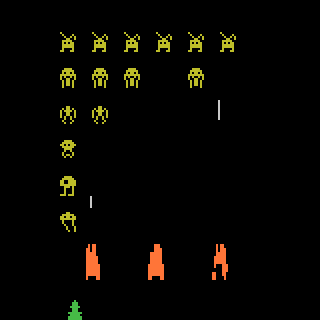}
    \includegraphics[width=.\fw\linewidth]{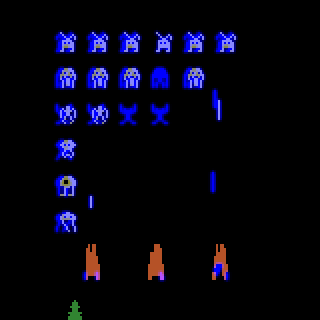}
    \includegraphics[width=.\fw\linewidth]{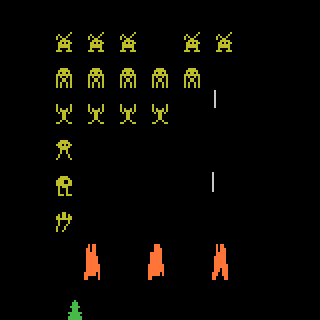}
    }
    \vspace{-2ex}
    \caption*{$a'=\text{MoveRight}$  \qquad  $a=\text{MoveLeftAndShoot}$,\qquad $a_{nn}'=\text{MoveLeft}$}
    \vspace{-2ex}
    \subfloat{
    \includegraphics[width=.\fw\linewidth]{images/user_study/generative_abl_agent/output3_0255_actionLEFTFIREr4_cf2r0RIGHT.png}
    \includegraphics[width=.\fw\linewidth]{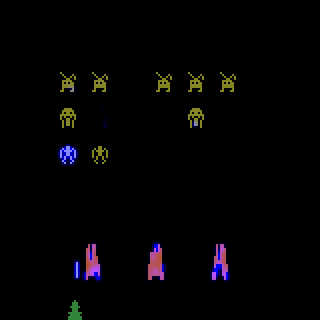}
    \includegraphics[width=.\fw\linewidth]{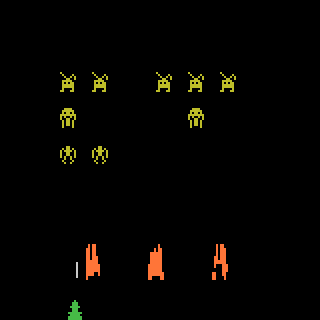}
    \includegraphics[width=.\fw\linewidth]{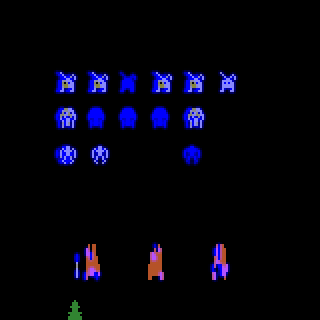}
    \includegraphics[width=.\fw\linewidth]{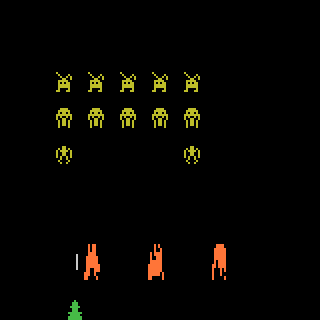}
    }
    \vspace{-2ex}
    \caption*{$a'=\text{MoveRight}$  \qquad  $a=\text{MoveLeftAndShoot}$,\qquad $a_{nn}'=\text{MoveLeft}$}
    \vspace{-2ex}
    \subfloat{
    \includegraphics[width=.\fw\linewidth]{images/user_study/generative_abl_agent/output3_0313_actionLEFTFIREr4_cf2r0RIGHT.png}
    \includegraphics[width=.\fw\linewidth]{images/user_study/generative_abl_agent/output2_0313_actionLEFTFIREr4_cf2r0RIGHT.png}
    \includegraphics[width=.\fw\linewidth]{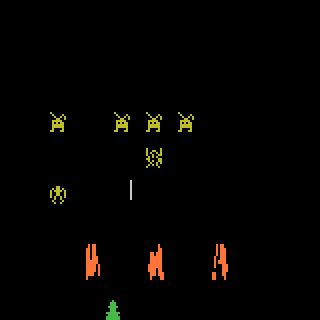}
    \includegraphics[width=.\fw\linewidth]{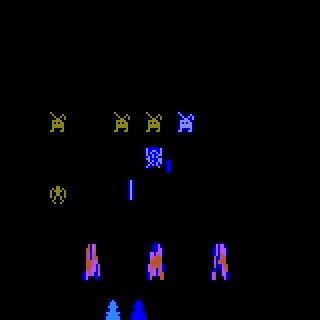}
    \includegraphics[width=.\fw\linewidth]{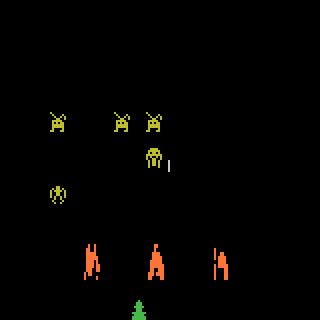}
    }
    \vspace{-2ex}
    \caption*{$a'=\text{MoveRight}$  \qquad  $a=\text{MoveLeftAndShoot}$,\qquad $a_{nn}'=\text{MoveLeft}$}
    \vspace{-2ex}

    \caption{
    Explanations 6 through 10 for the flawed agent used in the user study.
    (\textbf{Center}) The original state $s$ where the agent took action $a$.
    (\textbf{Left}) The counterfactual state explanations where the agent takes action $a'$.
    (\textbf{Right}) The Nearest Neighbor counterfactual state where the agent takes action $a_{nn}'$.
    (\textbf{Center Left/Right}) The highlighted difference between the counterfactual state and the original state.
         }

    \label{fig:space_allimgs_abl2}%
\end{figure}

\section{User study data analysis}
\label{sec:coding}


For answering research questions 2 and 3, two researchers collectively applied content analysis \citep{ContentAnalysisPaper2005} to the Post task questionnaire data corpus. They developed the codes shown in Table \ref{table:qualitative}. These codes were defined by having two researchers coded 20\% of the data corpus individually, achieving inter-rater reliability (IRR) of at least 90\% (calculated using Jaccard Index \citep{JaccardIndex}) with all the data sets.

\begin{table}[h]
\centering
\begin{tabular}{|p{1.6cm}|p{3.8cm}|p{5.3cm}|}
\hline
\textbf{Code} & \textbf{Description} & \textbf{Example} \\ \hline
Helpful & The participant found the artifact to be \textbf{helpful} to the main task, and it helped them better understand and evaluate the agent. & ``Yes the third image played a role in helping me make my decision.'' \\ \hline
Problematic & The participant found the artifact hinder-some and \textbf{problematic} in the main task. & ``The changed state portion confused me because I wasn't sure if that was the next action the AI took or the action it thought about taking given the highlighted circumstances.''\\ \hline
\end{tabular}
\caption{The qualitative codes used in our analysis}
 \label{table:qualitative}
\end{table}
